\documentclass[10pt,journal,compsoc]{IEEEtran}
\ifCLASSOPTIONcompsoc
  \usepackage[nocompress]{cite}
\else
  \usepackage{cite}
\fi
\ifCLASSINFOpdf
\else
\fi
\usepackage{graphicx}
\usepackage{booktabs}
\usepackage{multirow}
\usepackage{rotating}
\usepackage{wrapfig}
\usepackage{amssymb}
\usepackage{amsmath}
\usepackage{caption}
\usepackage{subcaption}
\usepackage{threeparttable}
\usepackage{hyperref}
\usepackage{placeins}
\usepackage{afterpage}
\usepackage{amsfonts}
\usepackage{booktabs}
\usepackage{multirow}
\usepackage{makecell}
\usepackage{stfloats}
\usepackage{titlesec}
\usepackage{array}   
\usepackage[table]{xcolor}

\titleformat{\paragraph}
{\normalfont\normalsize\bfseries}{\theparagraph}{1em}{}
\titlespacing*{\paragraph}
{0pt}{1.5ex plus 0.5ex minus .2ex}{1ex plus .2ex}
\usepackage{pifont}

\usepackage{tikz}

\usepackage[edges]{forest}
\usetikzlibrary{shadows.blur,shapes.multipart}
\tikzset{grimsel/.style={rectangle split,rectangle split parts=1,draw,
    fill=white,blur shadow,rounded corners,align=center}}
\begin{document}
%
\title{From Single-modal to Multi-modal Facial Deepfake Detection: Progress and  Challenges}
%
%
%
%

\author{Ping~Liu,~\IEEEmembership{Senior Member,~IEEE},
        Qiqi~Tao,~\IEEEmembership{}
        and~Joey Tianyi~Zhou,~\IEEEmembership{Senior Member,~IEEE}
\IEEEcompsocitemizethanks{\IEEEcompsocthanksitem P. Liu is with the Department of Computer Science and Engineering, University of Nevada, Reno, NV, 89512. 
 E-mail: pino.pingliu@gmail.com
\IEEEcompsocthanksitem Q. Tao and J. Zhou are with Centre for Frontier AI Research (CFAR), and  Institute of High Performance Computing (IHPC), Agency
for Science, Technology and Research(A*STAR), Singapore. Q. Tao and J. Zhou are also with Centre for Advanced Technologies in Online Safety (CATOS),  Singapore. \protect
E-mail: tao.qiqi@outlook.com; zhouty@cfar.a-star.edu.sg
\IEEEcompsocthanksitem J. Zhou is the corresponding author.
\IEEEcompsocthanksitem This work is done during Q. Tao's internship at Centre for Frontier AI Research (CFAR), A*STAR, Singapore.} \protect

}

\IEEEtitleabstractindextext{%
\begin{abstract}
As synthetic media, including video, audio, and text, become increasingly indistinguishable from real content, the risks of misinformation, identity fraud, and social manipulation escalate.
This survey traces the evolution of deepfake detection from early single-modal methods to sophisticated multi-modal approaches that integrate audio-visual and text-visual cues.
We present a structured taxonomy of detection techniques and analyze the transition from GAN-based to diffusion model-driven deepfakes, which introduce new challenges due to their heightened realism and robustness against detection.
Unlike prior surveys that primarily focus on single-modal detection or earlier deepfake techniques, this work provides the most comprehensive study to date, encompassing the latest advancements in multi-modal deepfake detection, generalization challenges, proactive defense mechanisms, and emerging datasets specifically designed to support new interpretability and reasoning tasks.
We further explore the role of Vision-Language Models (VLMs) and Multimodal Large Language Models (MLLMs) in strengthening detection robustness against increasingly sophisticated deepfake attacks.
By systematically categorizing existing methods and identifying emerging research directions, this survey serves as a foundation for future advancements in combating AI-generated facial forgeries.
{A curated list of all related papers can be found at \href{https://github.com/qiqitao77/Comprehensive-Advances-in-Deepfake-Detection-Spanning-Diverse-Modalities}{https://github.com/qiqitao77/Awesome-Comprehensive-Deepfake-Detection}.}
\end{abstract}

\begin{IEEEkeywords}
Generative Adversarial Network, Diffusion Model, Deepfake, Facial Forgery, Multi-modality
\end{IEEEkeywords}}

\maketitle

\IEEEdisplaynontitleabstractindextext

%
\IEEEpeerreviewmaketitle

\IEEEraisesectionheading{\section{Introduction}\label{sec:introduction}}

%
%
%
%
\IEEEPARstart{D}{eepfake}, encompassing video, audio, and text, has rapidly evolved due to advances in artificial intelligence. Early generative models, such as Variational Auto-Encoders (VAEs)\cite{kingma2019introduction} and Generative Adversarial Networks (GANs)\cite{goodfellow2014generative}, significantly improved the quality of synthetic media. 
More recently, Diffusion Models (DMs)~\cite{croitoru2023diffusion} have emerged as the dominant paradigm, offering higher visual fidelity, enhanced controllability, and greater robustness against detection. 
Unlike GANs, which rely on adversarial training, DMs progressively refine images through a noise-reduction process, making them less susceptible to conventional forgery artifacts. 
This shift has greatly increased the difficulty of distinguishing real content from synthetic forgeries, posing new challenges to human perception and deepfake detection systems.

Facial deepfakes particularly pose significant societal and ethical risks. While techniques like face swapping, attribute editing, and reenactment~\cite{pei2024deepfake} facilitate benign applications in entertainment and virtual avatars, they also enable malicious activities, including misinformation and financial fraud. 
Recent cases  demonstrated deepfake misuse in political disinformation\footnote{https://edition.cnn.com/2024/02/12/asia/suharto-deepfake-ai-scam-indonesia-election-hnk-intl/index.html} and corporate scams\footnote{https://edition.cnn.com/2024/02/04/asia/deepfake-cfo-scam-hong-kong-intl-hnk/index.html}, where AI-generated faces and voices have been used to manipulate public perception and deceive businesses.
As deepfake generation advances, these manipulations are no longer limited to a single modality (e.g., video or audio alone). 
The latest forgery techniques exploit multi-modal dependencies, including synchronizing fake speech with lip movements\cite{khalid2021fakeavceleb}, manipulating both video and textual descriptions\cite{shao2023detecting}, and simultaneously altering audio-visual content~\cite{cai2023av}. 
This trend toward multi-modal deepfakes has dramatically increased their realism, making traditional single-modal detection methods inadequate.

To counter these challenges, researchers have developed increasingly sophisticated detection strategies. 
Early deepfake detection focused on passive methods, which analyze content artifacts or inconsistencies {in fake content}. 
With the rise of multi-modal forgeries, recent approaches leverage deep learning models such as transformers~\cite{dosovitskiy2020image_iclr2021} and vision-language models~\cite{zhang2024vision_tpami2024} to improve detection capabilities. 
Various multi-modal fusion techniques\cite{zhou2021joint, khalid2021evaluation, raza2023multimodaltrace, wang2024avt2} have been proposed to enhance robustness by integrating visual, audio, and textual information. 
Additionally, detection efforts have expanded from passive detection to proactive solutions\cite{yu2020responsible_iclr2022, cui2023diffusionshield_nipsw2024}, {which disrupt unauthorized deepfake generation via adversarial perturbations or watermarking.}

Given the rapid evolution of deepfake techniques and detection strategies, this survey aims to provide the most comprehensive and up-to-date analysis of facial deepfake detection. 
As summarized in Table~\ref{tab:survey_comparison}, our work distinguishes itself from prior surveys in four key aspects:
\begin{itemize}
\item \textbf{Dedicated Focus on Facial Deepfake Detection.} 
Existing surveys~\cite{juefei2022countering, pei2024deepfake} relevant to deepfakes predominantly emphasize deepfake generation or general synthetic media detection, whereas our survey is exclusively tailored to the detection of facial deepfakes. 
Facial deepfakes have uniquely severe implications, affecting identity verification, public trust, and societal security. 
Therefore, our specialized focus provides an in-depth understanding essential for real-world applications and policy making.

\item \textbf{Early Comprehensive Coverage of Multi-modal Deepfake Detection.}
To the best of our knowledge, this survey was the first work to comprehensively cover multi-modal deepfake detection when initially completed.
While previous surveys~\cite{juefei2022countering, pei2024deepfake, wang2022gan_arxiv2022} limit their discussions primarily to visual manipulations, we systematically explore recent developments in detecting multi-modal forgeries that simultaneously involve audio, visual, and textual modalities. 
This work clearly categorizes these multi-modal approaches, identifying critical theoretical assumptions and practical limitations.

\item \textbf{Fine-grained and Structured Taxonomy of Detection Techniques.}
Unlike previous surveys that typically adopt broader categorizations~\cite{juefei2022countering, pei2024deepfake}, our taxonomy precisely distinguishes detection methods into finer hierarchical structures, such as passive vs. proactive approaches, single-modal vs. multi-modal detection, and nuanced categorizations based on underlying learning paradigms (e.g., consistency-based, learning-level innovations, and proactive interventions). 
This rigorous classification allows clearer insights into the strengths, limitations, and developmental trajectory of each technique category.

\item \textbf{Most Updated Coverage of Emerging Research (up to Mar 2025).}
Earlier surveys~\cite{juefei2022countering, wang2022gan_arxiv2022, wang2022deepfake, akhtar2023multimodal} predominantly address research published up to 2023, missing significant advancements in recent years. 
Our survey uniquely includes the latest breakthroughs, such as diffusion model-based manipulations, vision-language models, multi-modal large language models, and proactive defense strategies explicitly designed for diffusion-based techniques, thus providing readers with the most current and comprehensive view of this rapidly evolving field.





\end{itemize}
In summary, this survey provides a comprehensive and structured overview of facial deepfake detection, bridging the gap between single-modal and multi-modal approaches. 
By systematically categorizing existing techniques and highlighting emerging challenges, we offer a valuable resource that not only summarizes current progress but also identifies key directions for future research.

\textbf{Survey Structure:}
The remainder of this paper is organized as follows. Section~\ref{sec:background} provides background on deepfake detection, including key terminologies, datasets, and evaluation metrics. Section~\ref{sec:methods} systematically reviews detection methods, spanning single-modal to multi-modal and passive to proactive approaches (Figure~\ref{taxonomy}). Section~\ref{sec:future} discusses open challenges and future research directions. Finally, Section~\ref{sec:conclusion} summarizes key findings and insights.

\begin{figure*}[tb]
        \centering
        \includegraphics[width=0.8\textwidth]{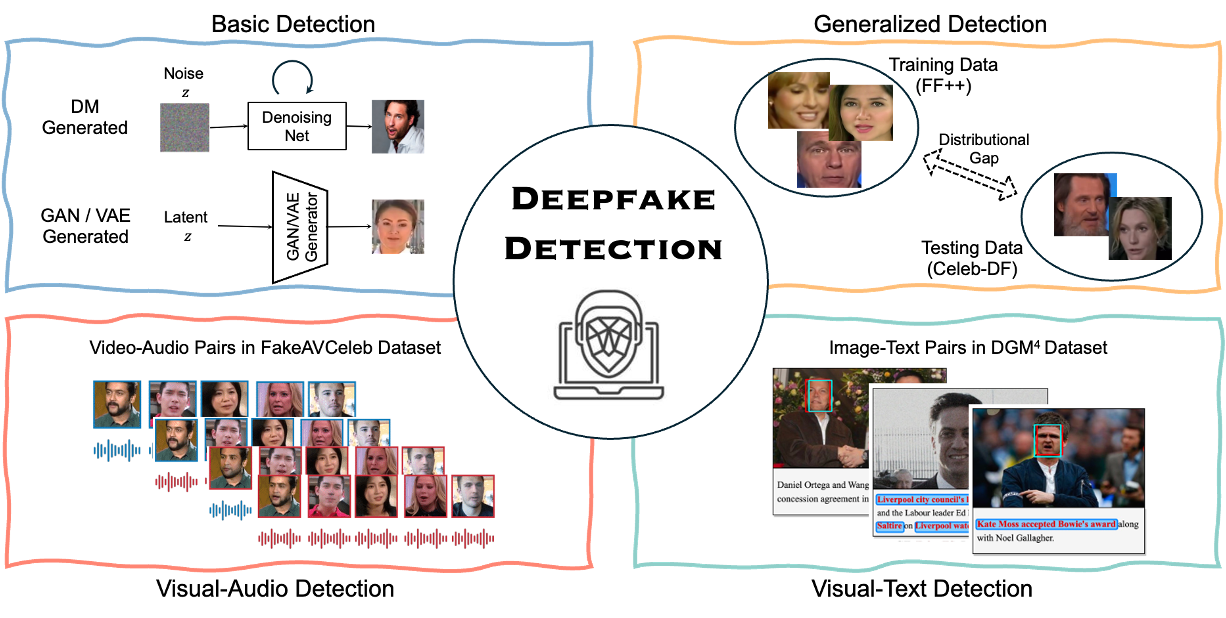}
        \caption{Advances in facial deepfake generation and corresponding detection in recent years.}
        \label{fig1}
\end{figure*}

\begin{table*}[htbp] 
    \caption{Comparison between this work and previous reviews. GAN: Generative Adversarial Network; VAE: Variational Auto-Encoder; DM: Diffusion Model.}
    \label{tab:survey_comparison}
    \setlength{\tabcolsep}{6pt} 
    \renewcommand{\arraystretch}{1.2} 
    \centering
    \begin{tabular}{c|c|c|c|c|c}
    \toprule
         \textbf{Work}& \textbf{Focused Tasks} & \textbf{Face-centric}& \textbf{Multi-modal}& \textbf{Generative Model}&\textbf{Released Date} \\
    \midrule
         Juefei-Xu et al.\cite{juefei2022countering}& Deepfake Generation and Detection& \checkmark &X & GAN \& VAE & Mar 2022\\
         Pei et al.\cite{pei2024deepfake}& Deepfake Generation and Detection & \checkmark& X& GAN \& VAE \& DM&Mar 2024 \\
         Lin et al.\cite{lin2024detecting}& AIGC Detection & X &\checkmark & LLM \& DM&  Feb 2024\\
         Wang et al.\cite{wang2022gan_arxiv2022} & Deepfake Detection  &\checkmark &X&GAN& Nov 2023 \\
         Wang et al.\cite{wang2022deepfake} & Reliability of Deepfake Detection&\checkmark & X & GAN \& VAE& Nov 2022\\
\rowcolor{orange!20}
\textbf{Ours} & \textbf{Facial Deepfake Detection}  & \textbf{\checkmark} & \textbf{\checkmark} & \textbf{GAN \& VAE \& DM} &  \textbf{Mar 2025} \\
    \bottomrule
    \end{tabular}
    
\end{table*}

\section{Background}
\label{sec:background}
\subsection{Task Formulation for Facial Forgery Detection}
In this subsection, we formally define the key tasks in facial deepfake detection, encompassing single-modal and multi-modal scenarios. 
Each task involves distinct input modalities and corresponding detection challenges.

{\noindent\textbf{Single-modal Detection.} 
Single-modal deepfake detection focuses on analyzing either visual or auditory input independently. 
A typical visual deepfake dataset is denoted as: $\mathcal{D} = \{(\mathbf{X_i^v}, Y_i)\}{i=1}^{N}, \quad \mathbf{X_i^v} \in \mathbb{R}^{T\times 3\times H\times W}, \quad Y_i \in \{real, fake\}$, where $\mathbf{X_i^v}$ represents a video with $T$ frames, and $Y_i$ indicates its authenticity. 
The detection model $F_\theta$ aims to classify real and fake samples: 

\begin{equation}
\theta = \operatorname*{arg\,min}\limits_{\theta{\prime}} \mathbb{E}_{\mathcal{D}} L(F_{\theta{\prime}}(\mathbf{X^v}), Y)
\end{equation}
Beyond binary classification, more advanced methods introduce spatial and temporal localization, where spatial localization predicts manipulated regions in an image or video frame, while temporal localization identifies manipulated segments across time.




{\noindent\textbf{Multi-modal Audio-Visual Detection}. 
As deepfake techniques advance, audio and visual content are often jointly manipulated to enhance realism, requiring detection models to analyze both modalities simultaneously \cite{zhou2021joint}. 
Given an audio-visual dataset: $\mathcal{D} = \{(\mathbf{X_i^v}, \mathbf{X_i^a}, Y)\}_{i=1}^{N}$, where $\mathbf{X_i^a} \in \mathbb{R}^{M \times c}$ represents the audio waveform with $c$ channels and $M$ sample points, the detection function is defined as: 
\begin{equation}
    \theta = \operatorname*{arg\,min}\limits_{\theta{\prime}} \mathbb{E}_{\mathcal{D}} L(F_{\theta{\prime}}(\mathbf{X^v}, \mathbf{X^a}), Y)
\end{equation}
Additionally, fine-grained classification may be required, where $F_\theta$ separately predicts the authenticity of the visual and audio components, i.e., $Y^v\in\{real, fake\}$ and $Y^a\in\{real, fake\}$, to identify  forgeries in respective modalities.




\noindent\textbf{Multi-modal Text-Visual Detection and Grounding}
Recently, deepfake content extends beyond audio-visual manipulations, with text-visual inconsistencies becoming increasingly common (e.g., altered captions or misleading news images). 
A text-visual dataset is represented as: $\mathcal{D} = \{(\mathbf{X_i^v}, \mathbf{X_i^t}, Y^v_i, Y^t_i, \mathbf{B^v_i}, \mathbf{M^t_i}, Y_i)\}_{i=1}^{N}$, where $\mathbf{X^t_i} \in \mathbb{R}^{L \times d}$ is a textual description, and $\mathbf{B^v}$ and $\mathbf{M^t}$ denote the manipulated regions in image and text, respectively. 
In this setting, the detector learns to minimize the designed loss: 
{
\begin{equation}
    \theta = \operatorname*{arg\,min}\limits_{\theta'} \mathbb{E}_{\mathcal{D}} \\
    L(F_{\theta'}(\mathbf{X^v},\mathbf{X^t}),Y,Y^v,Y^t,\mathbf{B^v},\mathbf{M^t}).
\end{equation}
}

{\subsection{Datasets for Facial Forgery Detection}}
The development and evaluation of deepfake detection algorithms heavily rely on diverse datasets, which can be categorized as single-modal or multi-modal. 
Single-modal datasets focus on one type of manipulated content, such as images or audio, facilitating targeted detection of visual or auditory artifacts. 
In contrast, multi-modal datasets incorporate multiple data streams (e.g., visual, audio, and text), reflecting real-world deepfake scenarios where different modalities are simultaneously manipulated. 
These datasets enable more robust detection by leveraging cross-modal inconsistencies, such as lip-sync mismatches or image-text semantic conflicts. 
However, challenges such as data bias, limited diversity, and real-world generalization issues remain. Table \ref{table:benchmark} provides a detailed comparison of publicly available datasets for facial forgery detection.

\begin{table*}[htbp]
  \caption{Overview of Deepfake Datasets focusing on facial forgery.  A: audio modality; V: visual modality; T: text modality; BC: binary classification; TG: temporal grounding; SG: spatial grounding; IN: interpretation; FS: face swap; FR: face reenactment; {FE: face editing}; {TTS: text-to-speech}; VC: voice conversion; WIG: whole image generation; {TS: text swap}.} 
  \label{table:benchmark}
    \centering
    \begin{threeparttable}
    \begin{tabular}{@{}lcc>{\centering\arraybackslash}p{1cm}>{\centering\arraybackslash}p{1cm}>{\centering\arraybackslash}p{1cm}cccc@{}}
    \toprule
    \textbf{Dataset\tnote{\footnotesize *}} & \textbf{Year} & \textbf{Type} & \textbf{Tasks} & \multicolumn{3}{c}{\textbf{Manipulation Method}} &\textbf{\#Subjects} & \textbf{\#Real} & \textbf{\#Fake} \\
    \cmidrule(lr){5-7} \cmidrule(lr){8-10}
    & & & & \textbf{A} & \textbf{V} & \textbf{T} & & & \\
    \midrule
\rowcolor{green!40}    FaceForensics++ \cite{rossler2019faceforensics} & 2019 & Visual, GAN/VAE& BC & - & FS/FR & - & - & $1,000$ & $4,000$ \\
   \rowcolor{green!40} DFD \cite{dfd} & 2019 & & BC & - & FS/FR & - & - & $363$ & $3,068$ \\
 \rowcolor{green!40}   FaceShifter\cite{li2020faceshifter} & 2020 & & BC & - & FS & - & - & - & $1,000$ \\
\rowcolor{green!40}DFDC~\cite{dolhansky2020deepfake} & 2020& &  BC &  - & {FS} &  {-} & {$960$}  & {$23,654$} & {$104,500$}\\
\rowcolor{green!40}{Celeb-DF\cite{li2020celeb}} & {2020}& & {BC} &   {-} &{FS} &  {-} & {$59$}  & {$590$} & {$5,639$}\\

 \rowcolor{green!40}{DeeperForensics-1.0\cite{jiang2020deeperforensics}} &{2020}& &  {BC} &  {-} & {FS} &  {-} & {$100$}  & {$50,000$} & {$10,000$}\\
 \rowcolor{green!40}{WildDeepfake\cite{zi2020wilddeepfake}} &{2020} & &  {BC}  & {-} & {-} &  {-} & {-}  & {$3,805$} & {$3,509$}\\
 \rowcolor{green!40}{KoDF\cite{kwon2021kodf}} &{2020} & &  {BC} &  {-} & {FS} &  {-} & {{$403$}}  & {{$62,166$}} & {{$175,776$}}\\

\rowcolor{green!40} {FFIW$_{10K}$\cite{zhou2021face}} &{2021} & &  {BC/SG} & {-} & {FS} &  {-} & {-}  & {$10,000$} & {$10,000$}\\
 \rowcolor{green!40}{ForgeryNet\cite{he2021forgerynet}} &{2021} & &  {BC/SG/TG} &  {-} & {FS/FR} &  {-} & {$5,400$}  & {$99,630$} & {$121,617$}\\
\rowcolor{green!40} {DF-Platter\cite{narayan2023df}} &{2023} & &  {BC} & {-} & {FS} &  {-} & {$454$}  & {$133,260$} & {$132,496$}\\
\midrule
\rowcolor{green!20}{DeepFakeFace\cite{song2023robustness}} &{2023} & Visual, DM &  {BC} &{-} &{WIG} & {-} &  {-}  & {$30,000$} & {$90,000$}\\
\rowcolor{green!20}{DiffusionFace \cite{chen2024diffusionface}} &{2024}& &  {BC}  &{-} &{FS/WIG} & {-} & {-}  & {$30,000$} & {$600,000$}\\

\rowcolor{green!20}{DiFF \cite{cheng2024diffusion_arxiv2024}} &{2024} & &  {BC} &{-} &{WIG/FS} & {-} & {$1,070$}  & {{$23,661$}} & {{$537,466$}}\\
\rowcolor{green!20}{VLF \cite{he2025vlforgeryfacetriaddetection_arxiv2025}} &{2025} & &  {BC/SG} &{-} &{WIG/FS} & {-} & {-}  & {{$96.5K$}} & {{$445.6K$}}\\
\midrule
\rowcolor{green!20}{MMTT \cite{lian2024large_arxiv2024}} &{2024} & Visual, GAN/DM&  {BC/IN} &{-} &{FS/FR} & {-} & {-}  & {{-}} & {{-}}\\ 
\rowcolor{green!20}{ExDDV \cite{hondru2025exddvnewdatasetexplainable_arxiv2025}} &{2025} & Video, GAN&  {BC/IN} &{-} &{FS/FR} & {-} & {-}  & {{1,000}} & {{4,369}}\\ 
\midrule
\midrule
\rowcolor{blue!20}{FakeAVCeleb \cite{khalid2021fakeavceleb}} &{2021} & {Audio-Visual} &  {BC}  &{SV2TTS} &{FR} & {-} & {{$500$}}  & {{$500$}} & {{$19,500$}}\\
\rowcolor{blue!20}{LAV-DF \cite{10034605}} &{2022} &  &  {BC/TG}  &{TTS} &{FR} & {-} & {$153$}  & {$36,431$} & {$99,873$}\\
\rowcolor{blue!20}{AV-Deepfake1M\cite{cai2023av}} &{2023} &\multirow{3}{*}{Audio-Visual Deepfake} &  {BC/TG}&{TTS} &{FR} & {-} & {$2,068$}  & {$286,721$} & {$860,039$}\\

\rowcolor{blue!20}{PolyGlotFake\cite{hou2024polyglotfake_arxiv2024}} &{2024} & &  {BC} &{TTS/VC} &{FR} & {-} & {-}  & {$766$} & {$14,472$}\\

\rowcolor{blue!20}{DeepFake-eval-2024\cite{chandra2025deepfake_arxiv2025}} &{2025} & &  {BC} &\multicolumn{3}{c}{Collected from the Wild} & {-}  & {$1208$} & {$767$}\\

\midrule
\midrule
\rowcolor{orange!20}{DGM$^4$\cite{shao2023detecting}} &{2023} & Text-Visual &  {BC/SG} &{-} &{FS/FE} & {TS} & {-}  & {$77,426$} & {$152,574$}\\
    \bottomrule
    \end{tabular}
    
\begin{tablenotes}
    \item[*]Only publicly available datasets are listed.
\end{tablenotes}
  
    \end{threeparttable}
\end{table*}

\subsubsection{Single-modal (visual) Deepfake Datasets}
The evolution of single-modal deepfake datasets has progressed from traditional computer graphics-based techniques to generative AI methods such as GANs, VAEs, and DMs~\cite{rossler2019faceforensics, li2020celeb, jiang2020deeperforensics, dolhansky2020deepfake, zi2020wilddeepfake, kwon2021kodf, chen2024diffusionface, song2023robustness}. 
These datasets serve as fundamental benchmarks for training and evaluating deepfake detection algorithms, yet they also face challenges such as limited generalization across manipulation techniques and demographic biases.

\paragraph{GANs/VAEs based Generated Datasets.} Early datasets primarily utilized GANs and VAEs to generate manipulated facial images and videos, establishing a foundation for deepfake detection. 
FaceForensics++ (FF++) \cite{rossler2019faceforensics} introduced multiple manipulation techniques, including FaceSwap, Face2Face, DeepFakes, and NeuralTextures, setting a benchmark for detection research.
DFD \cite{dfd} further contributed by providing both pristine and manipulated videos featuring paid, consenting actors, ensuring high-quality ground-truth data. 
DFDC \cite{dolhansky2020deepfake} extended this by integrating diverse forgery methods, such as FSGAN \cite{nirkin2019fsgan} and StyleGAN \cite{karras2019style}. 
Celeb-DF \cite{li2020celeb} improved upon earlier datasets by using higher-quality face-swapping techniques to reduce detectable artifacts, while DeeperForensics-1.0 \cite{jiang2020deeperforensics} introduced DF-VAE, a learning-based framework for more natural face reenactment.

\noindent \textbf{Datasets for Spatial and Temporal Localization.}
To expand detection challenges beyond binary classification, datasets such as FFIW$_{10K}$ \cite{zhou2021face} and ForgeryNet \cite{he2021forgerynet} introduced spatial and temporal annotations. 
FFIW$_{10K}$ \cite{zhou2021face} focuses on multi-person forgery detection, requiring models to distinguish manipulated individuals within a group. 
ForgeryNet \cite{he2021forgerynet}, covering $5,400$ subjects and $15$ manipulation techniques, supports spatial and temporal localization tasks.

\noindent \textbf{Demographic Diversity in Deepfake Datasets.}
Recognizing dataset biases, recent efforts have emphasized improving subject diversity. 
KoDF\cite{kwon2021kodf} targets underrepresented Asian demographics, incorporating multiple synthesis models to enhance realism. 
DF-Platter\cite{narayan2023df} focuses on Indian ethnicity and gender balance, ensuring more equitable representation across demographic groups. 
Such datasets help mitigate performance disparities in deepfake detection across different populations.

\noindent \textbf{Real-world Scenario Testing.}
Traditional datasets primarily generate synthetic manipulations, but real-world deepfakes differ due to compression artifacts, post-processing, and lower-quality forgeries. 
WildDeepfake\cite{zi2020wilddeepfake} directly collects high-quality deepfake samples from online sources rather than artificially creating them, serving as a realistic benchmark for deployed detection models. 
Furthermore, during the COVID-19 pandemic, researchers\cite{ko2022deepfake_wdc2022} introduced a masked deepfake dataset, revealing that face occlusion significantly reduces detection accuracy, highlighting the importance of robustness testing in practical conditions.

\paragraph{DMs based Generated Datasets.} 
Recent advances in Diffusion Models (DMs) have driven significant progress in deepfake dataset generation, leading to datasets such as DeepFakeFace \cite{song2023robustness}, DiffusionFace \cite{chen2024diffusionface}, DiffusionDB-Face/JourneyDB-Face \cite{Bhattacharyya2024DiffusionD}, and DiFF \cite{cheng2024diffusion_arxiv2024}. 

Specifically, DeepFakeFace \cite{song2023robustness}employs three distinct diffusion methods (Stable Diffusion v1.5, Stable Diffusion Inpainting, and InsightFace) to generate $90,000$ diverse synthetic facial images, offering substantial data diversity. 
DiffusionFace \cite{chen2024diffusionface}, the first dedicated dataset for diffusion-based facial forgery detection, utilizes $11$ DMs covering manipulations from unconditional generation to face swapping, thereby establishing a comprehensive benchmark. 
DiffusionDB-Face/JourneyDB-Face \cite{Bhattacharyya2024DiffusionD} provide advanced benchmarks emphasizing variations in head poses, facial attributes, and artistic styles, while the extensive DiFF dataset \cite{cheng2024diffusion_arxiv2024} includes over $500,000$ images synthesized using $13$ different DM-based methods with carefully curated textual and visual prompts for semantic and visual coherence.
Recently, Yan et al.introduced DF40\cite{df40_arXiv2024}, a large-scale dataset incorporating $40$ distinct deepfake techniques spanning GAN- and DM-based methods. 
Through comprehensive evaluations, including experiments $2,000$ with four standard protocols and seven detection methods, DF40 highlighted key insights, including CLIP-large’s superior generalization capabilities across face and non-face deepfakes.

Extending beyond traditional face manipulation, recent diffusion-based datasets have also targeted broader deepfake contexts, emphasizing realistic dissemination scenarios, diverse application domains, and critical societal issues such as fairness and bias.
RWDF-23\cite{cho2023towards_cikm2023} comprises 2,000 deepfake videos from multiple online platforms in four languages, accompanied by viewer interactions and context analysis, reflecting realistic deepfake dissemination. Zoom-DF\cite{park2022zoom_wsidc2022} uniquely targets manipulations specific to remote meetings and video conferencing, focusing on altered participant movements rather than traditional face swapping. 
Additionally, AI-Face~\cite{lin2024ai_arxiv2024} addresses fairness and bias in deepfake detection by providing a million-scale dataset of demographically annotated, 
generated facial images, highlighting demographic representation as a critical factor in evaluation.

\paragraph{Interpretability-oriented Benchmark}
Given the increasing demand for datasets explicitly supporting interpretability in deepfake detection, Lian et al.~\cite{lian2024large_arxiv2024} introduced MMTT, a large-scale interpretability-focused benchmark containing over $128,303$ image-text pairs generated using diverse GANs and diffusion models. 
In MMTT, each sample is manually annotated with structured textual descriptions and pixel-level binary masks, explicitly facilitating fine-grained reasoning and precise localization of manipulated regions.
{Moving beyond broad coverage, He et al.~\cite{he2025vlforgeryfacetriaddetection_arxiv2025} introduced the VLF dataset, distinguished by its fine-grained, chain-of-thought annotations that integrate external generation knowledge and low-level visual comparisons, enhancing the interpretability of advanced detection and boosting localization, and attribution tasks.}
To extend interpretability research beyond static images to video-based deepfake detection, Hondru et al.~\cite{hondru2025exddvnewdatasetexplainable_arxiv2025} introduce ExDDV, the first dataset and benchmark for explainable deepfake detection in video. 
ExDDV comprises $5.4$K real and deepfake videos, each annotated with text descriptions explaining artifacts and clicks marking their locations, enabling the development of detection models that not only detect but also explain deepfake content at a fine-grained level.

\begin{figure*}[htbp]
  \centering
      \begin{subfigure}{0.45\textwidth}
    \includegraphics[width=\textwidth]{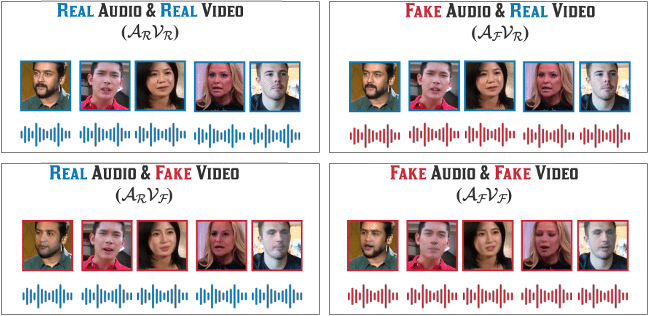}
      \caption{Multi-modal audio-visual deepfake dataset FakeAVCeleb\cite{khalid2021fakeavceleb}, figure modified from \cite{khalid2021fakeavceleb}.}
      \label{multi-modal-av}
  \end{subfigure}
  \hfill
    \begin{subfigure}{0.5\textwidth}
    \includegraphics[width=\textwidth]{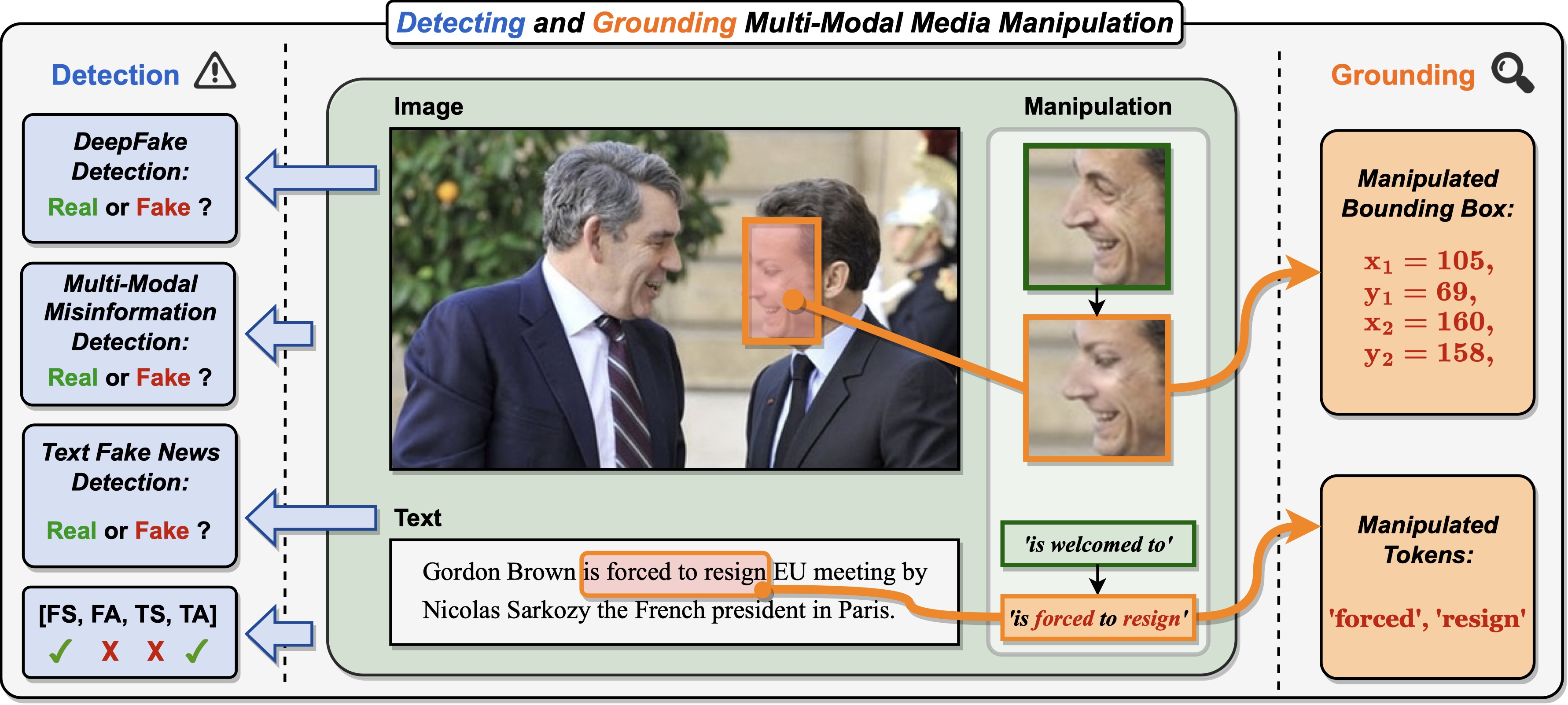}
      \caption{Multi-modal visual-text deepfake dataset $DGM4$, figure from \cite{shao2023detecting}.}
      \label{multi-modal-vt}
  \end{subfigure}
  \caption{Multi-modal deepfake detection.}
    \vspace{-0.3cm}
    \label{multi-modal-datasets}
\end{figure*}

\subsubsection{Multi-modal Deepfake Datasets}
Multi-modal deepfake datasets are essential for realistically evaluating detection algorithms, as they mirror real-world scenarios where misinformation involves combined manipulations of videos, audio, and textual content. 
Such datasets enhance the detection robustness by providing complex cross-modal inconsistencies. 
As depicted in Figure~\ref{multi-modal-datasets}, multi-modal datasets typically comprise audio-visual manipulations, where video and audio tracks are forged, or text-visual manipulations, involving falsified text-image pairs.

\paragraph{Audio-Visual Deepfake Datasets}
Audio-visual deepfake datasets integrate synthesized speech with manipulated visual content, addressing challenges such as lip synchronization and auditory realism. 
Datasets like FakeAVCeleb\cite{khalid2021fakeavceleb} and Joint Audio-Video Deepfake\cite{zhou2021joint} significantly contribute to this domain by incorporating realistic audio-visual manipulations. 
The Trusted Media Challenge (TMC)\cite{chen2022trusted} and DefakeAVMiT\cite{10081373} datasets further advance detection capabilities by including diverse audio-visual forgeries. 
Additionally, PolyGlotFake\cite{hou2024polyglotfake_arxiv2024}, the first multilingual audio-visual deepfake dataset, spans seven languages, enriching the linguistic diversity crucial for robust detection methods. 
Specialized datasets such as Localized Audio Visual DeepFake (LAV-DF)\cite{10034605} and AV-Deepfake1M~\cite{cai2023av} specifically target temporal forgery localization, reflecting the increasing subtlety and sophistication of audio-visual manipulations. 
{The well-curated academic datasets are limited to content diversity and contain human differentiable artifacts, resulting in their inability to mirror the complicated deepfakes in a realistic situation. DeepFake-Eval-2024\cite{chandra2025deepfake_arxiv2025} addresses this limitation by providing in-the-wild deepfakes collected from social media and various deepfake detection platforms, thereby becoming a challenging testbed for deepfake detection tools.}

\paragraph{Text-Visual Deepfake Datasets}
Text-visual datasets evaluate detectors against image-text inconsistencies or manipulation across modalities. 
The DGM4 dataset~\cite{shao2023detecting} represents a notable advancement in this area, focusing specifically on facial forgeries within multimodal contexts. 
Unlike traditional text-image mismatch datasets, DGM4 challenges models to simultaneously analyze visual and textual content, supporting not only binary classification but also detailed manipulation localization within images and corresponding textual descriptions. 
Such complexity emphasizes the growing importance of comprehensive multimodal analysis to effectively counter sophisticated multimedia forgeries.

\subsection{Evaluation}
\noindent\textbf{Evaluation Metrics}
Evaluation metrics in deepfake detection depend on both the task formulation and the specific modality involved, and can be summarized into the following categories:
\begin{itemize}
    \item \textit{Binary classification}: This formulation is typically used in single-modal (visual-only) detection scenarios, employing common metrics such as Accuracy (ACC), Area Under ROC Curve (AUC), F1-score, Average Precision (AP), and Equal Error Rate (EER).

    \item \textit{Multi-label classification}: This task is applied in multi-modal detection settings that involve fine-grained classification across multiple modalities~\cite{raza2023multimodaltrace,shao2023detecting}. Here, Mean Average Precision (mAP) is additionally adopted for comprehensive performance evaluation.

    \item \textit{Grounding and localization}: This category is utilized in both single-modal and multi-modal tasks, measuring spatial localization accuracy through metrics such as mean Intersection over Union (mIoU), IoU$_{50}$, and IoU$_{75}$, which evaluate bounding box predictions for manipulated regions. For tasks involving textual grounding, token-level precision and recall are further reported to quantify textual localization accuracy~\cite{shao2023detecting}.
\end{itemize}

\noindent\textbf{Evaluation Settings} 
Evaluation protocols for deepfake detection are generally categorized into \textit{in-domain} and \textit{cross-domain} settings, based on the distributional alignment between training and testing data.

\begin{itemize}
    \item \textbf{In-domain evaluation} involves testing detectors on data from the same distribution as the training set. 
    Previous works~\cite{rossler2019faceforensics,dolhansky2020deepfake,jiang2020deeperforensics} typically evaluate models using separate, non-overlapping subsets of the same dataset or identical manipulation techniques. 
    The results of the in-domain evaluations reflect the detection performance under controlled conditions, offering insights into the effectiveness of the algorithm within idealized scenarios.

    \item \textbf{Cross-domain evaluation}, conversely, assesses detectors' ability to generalize when training and test data distributions differ, closely mimicking real-world scenarios. 
    Three primary cross-domain evaluation settings exist:
    \begin{itemize}
        \item \textit{Cross-dataset:} Training and testing datasets differ (e.g., training on FF++~\cite{rossler2019faceforensics}, testing on Celeb-DF~\cite{li2020celeb}).
        
        \item \textit{Cross-manipulation:} Using the same dataset but training and testing on different manipulation techniques (e.g., training on FaceSwap, testing on NeuralTextures within FF++).

        \item \textit{Cross-postprocessing:} Testing detectors on data subjected to post-processing operations (e.g., JPEG or video compression~\cite{jiang2020deeperforensics}), simulating practical scenarios where media content undergoes compression or other alterations before dissemination. 
        For example, models might be trained on high-quality videos from FF++ but evaluated on compressed versions, assessing robustness to realistic media processing conditions.
    \end{itemize}
\end{itemize}

\subsection{History and Challenges of Deepfake Datasets}
The evolution of deepfake datasets has closely followed advances in generative AI, transitioning from early GAN and VAE-based techniques~\cite{rossler2019faceforensics,dolhansky2020deepfake,dfd,li2020celeb} to sophisticated Diffusion Models (DM)~\cite{song2023robustness,chen2024diffusionface,cheng2024diffusion_arxiv2024}. 
This progression has necessitated significant shifts in detection strategies, from identifying obvious visual artifacts to decoding subtle patterns that closely mimic genuine content, posing greater challenges to both machines and humans.

Moreover, the recent emergence of multi-modal deepfake datasets~\cite{khalid2021fakeavceleb,10034605,cai2023av,hou2024polyglotfake_arxiv2024,shao2023detecting} further complicates detection tasks. 
Detectors must now analyze cross-modal interactions, reflecting realistic scenarios in which misinformation integrates multiple manipulated data streams, including video, audio, and textual content.
Additionally, manipulation techniques within deepfakes have grown increasingly subtle and localized, targeting specific facial regions in images~\cite{zhou2021face,he2021forgerynet,shao2023detecting} or selected frames within video sequences~\cite{10034605,cai2023av}. 
Such localized and temporally selective manipulations require that detection systems not only classify content authenticity but also precisely pinpoint manipulated regions and temporal segments.

In summary, these advances across generative methods, modality integration, and increasingly sophisticated manipulations highlight persistent challenges in dataset design and emphasize the necessity for continuously adaptive and robust deepfake detection methodologies.

\section{Method} \label{sec:methods}
In this section, we systematically review recent advances in deepfake detection methods published primarily over the past three years, categorizing them into single-modal and multi-modal approaches. 
Due to the breadth and technical complexity of available techniques, we focus on key representative works within each category, aiming to clearly illustrate the evolution and significant methodological milestones. 
Figure~\ref{taxonomy} summarizes our detailed taxonomy for single-modal and multi-modal deepfake detection methods.

    \tikzset{
  non leaf/.style={
    draw=gray,
    thick,
    minimum height=1ex,
    rounded corners=3,
    text width=2.1cm,
    align=center,
    font=\sffamily\scriptsize,
    text centered,
  },
  leaf/.style={
      draw=gray,
    thick,
    minimum height=1ex,
    rounded corners=3,
    text width=2.6cm,
    align=center,
    font=\sffamily\tiny,
    text centered  }
}
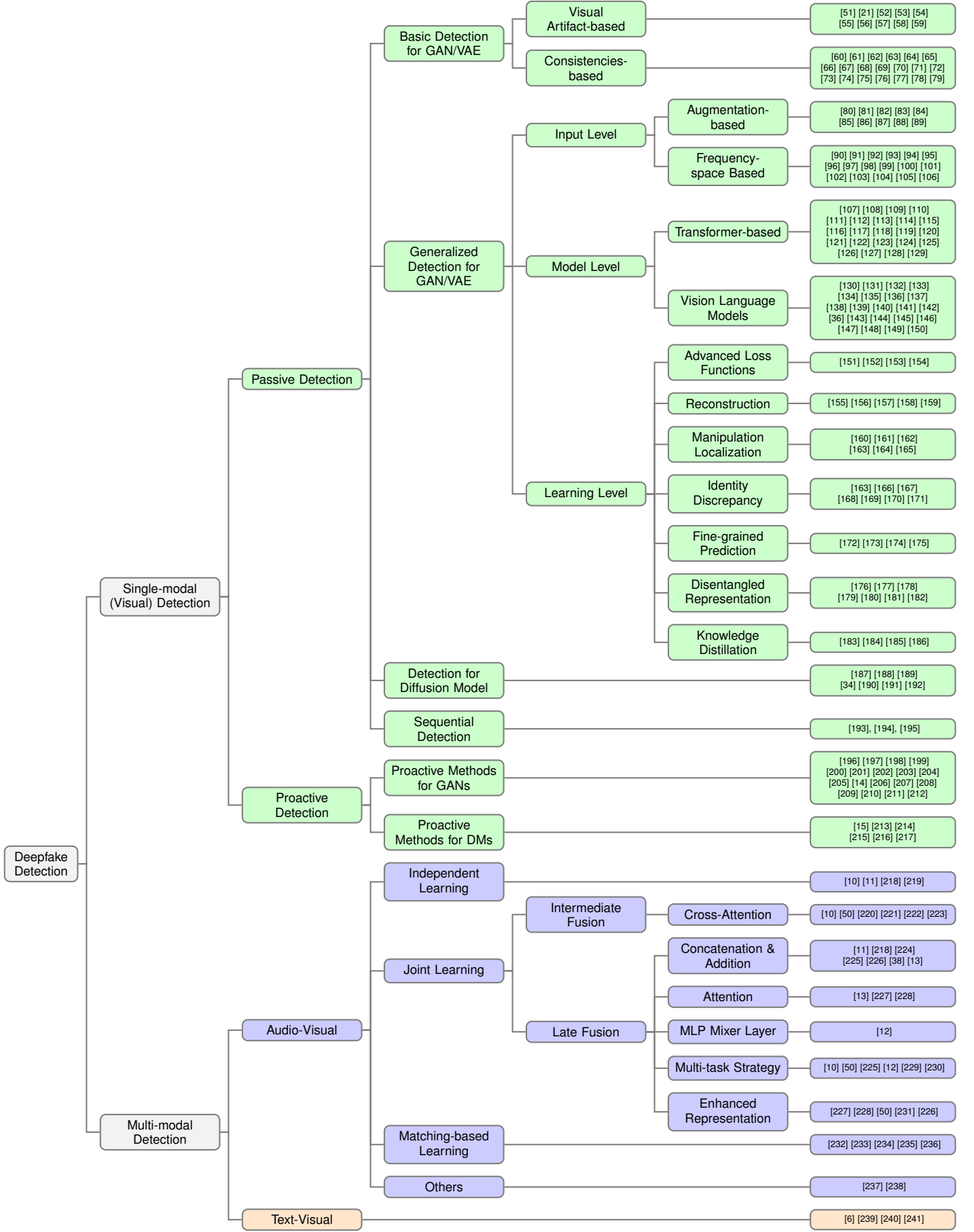
\begin{figure*}[htbp]
    \centering
    \hspace{-0.5cm}
    \begin{forest}
    for tree={
    where n children=0{leaf}{non leaf},
    l sep+=0.5pt,
    grow'=east,
    edge={gray, thick},
    parent anchor=east,
    child anchor=west,
    calign=center,
    calign primary child=1,
    edge path={
      \noexpand\path [draw, \forestoption{edge}] (!u.parent anchor) -- +(4.5pt,0) |- (.child anchor)\forestoption{edge label, s sep=10pt};
    },
    if={isodd(n_children())}{
      for children={
        if={equal(n,(n_children("!u")+1)/2)}{calign with current}{}
      }
    }{},
  }
  [Deepfake Detection, text width=1.2cm,fill=gray!10
    [Single-modal (Visual) Detection, tier=level1, fill=gray!10
      [Passive Detection, tier=level2, fill=green!20
        [Basic Detection for GAN/VAE, tier=level3, fill=green!20
            [Visual Artifact-based, tier=level4, calign=child edge, calign child=1, fill=green!20
                [\cite{afchar2018mesonet}\cite{rossler2019faceforensics}\cite{liu2020global_cvpr2020}\cite{zhao2021multi}\cite{tan2023rethinking_cvpr2024}\cite{tan2024data}\cite{corvi2023intriguing_cvprw2023}\cite{chen2024single_arxiv2024}\cite{10246417_tmm2023}\cite{nguyen2024laa_cvpr2024}, tier=last, fill=green!20]
            ]
            [Consistencies-based, tier=level4, calign=child edge, calign child=1, fill=green!20
                [\cite{8630787}\cite{qi2020deeprhythm_mm2020}\cite{zheng2021exploring_iccv2021}\cite{haliassos2021lips}\cite{choi2024exploiting_cvpr2024}\cite{chu2025reducedspatialdependencygeneral_icassp2025}\cite{latentpattern_acmmma2022}\cite{haliassos2022leveraging_cvpr2022}\cite{pang2023mre_tcsvt2023}\cite{wang2023altfreezing}\cite{chen2025gc_arxiv2025}\cite{fang2024uniforensics_arxiv2024}\cite{nguyen2025vulnerability_arxiv2025}\cite{zhangnaco_arxiv2024}\cite{xu2023tall_iccv2023}\cite{xu2024towards}\cite{liu2023ti2net_wacv2023}\cite{wang2023noise_aaai2023}\cite{chen2024compressed_arxiv2024}\cite{yan2024generalizing_arxiv2024}, tier=last, fill=green!20]
            ]
        ]
        [Generalized Detection for GAN/VAE, tier=level3, fill=green!20, calign=child edge, calign child=2
            [Input Level, tier=level4, fill=green!20
                [Augmentation-based, tier=level5, calign=child edge, calign child=1, fill=green!20
                    [\cite{Wang2021_cvpr2021}\cite{zhao2021learning_iccv2021}\cite{chen2022self_cvpr2022}\cite{shiohara2022detecting_cvpr2022}\cite{larue2023seeable_iccv2023}\cite{li2023pixel_tmm2023}\cite{cheng2024can_arxiv2024}\cite{lin2024faketillmakeit_eccv2024}\cite{stamnas2025difffake_wacvw2025}\cite{ahmed_arxiv2024}, tier=last, fill=green!20]
                ]
                [Frequency-space Based, tier=level5, calign=child edge, calign child=1, fill=green!20
                    [\cite{qian2020thinking}\cite{chen2021local}\cite{li2021frequency}\cite{yan2023transcending}\cite{li2024freqblender_arxiv2024}\cite{song2024quality_arxiv2024}\cite{luo2021generalizing}\cite{liu2021spatial}\cite{woo2022add}\cite{jeong2022frepgan}\cite{wang2023dynamic_cvpr2023}\cite{tan2024frequency}\cite{doloriel2024frequency_icassp2024}\cite{li2024multiple_arxiv2024}\cite{baru2024harnessing_arxiv2024}\cite{peng2025wmamba_arxiv2025}\cite{yan2025generalizable_arxiv2025}, tier=last, fill=green!20]
                ]
            ]
            [Model Level, tier=level4, fill=green!20
                [Transformer-based, tier=level5, calign=child edge, calign child=1, fill=green!20
                    [\cite{Wang2024ATS_arxiv2024}\cite{saha2023undercover_iccvw2023}\cite{bai2023aunet_cvpr2023}\cite{zhang2023ummaformer_acmmm2023}\cite{pellicer2024pudd_cvprw2024}\cite{tsai2024understanding_arxiv2024}\cite{zhang2024genface_arxiv2024}\cite{fu2025exploring_aaai2025}\cite{chen2024guided_arxiv2024} \cite{guo2024face_arxiv2024}\cite{li2025deepfakedetectionknowledgeinjection_arxiv2025} \cite{nguyen2024fakeformer_arxiv2024} \cite{luo2024forgery_tcsvt2024} \cite{zhang2025distilled_pr2025} \cite{Kong2024OpenSetDD_arxiv2024}\cite{shao2023deepfake_arxiv2023}\cite{cui2024forensics_arxiv2024}\cite{liu2024mixture_arxiv2024}\cite{kong2024moe_arxiv2024}\cite{lin2024standing_aaai2025}\cite{guo2025rethinkingvisionlanguagemodelface_arxiv2025}\cite{kundu2024towards_arxiv2024} \cite{nie2024dip_tmm2024}, tier=last, fill=green!20]
                ]
                [Vision Language Models, tier=level5, calign=child edge, calign child=1, fill=green!20
                    [\cite{han2024towards_arxiv2024}\cite{Lai2024GMDFGM_arxiv2024}\cite{tan2024c2pclip_arxiv2024}\cite{lin2024standingshouldersgiantsreprogramming_arxiv2024}\cite{wang2025knowledge_arxiv2025}\cite{kong2023enhancing_wacv2025}\cite{zhang2024mfclipmultimodalfinegrainedclip_arXiv2024}\cite{zhang2024dpl_accv2024}\cite{levy2024nearly_arxiv2024}\cite{tsigos2024towards_icmrw2024}\cite{sun2023towards_arxiv2023}\cite{zhang2024common_arxiv2024}\cite{huang2024ffaa_arxiv2024}\cite{lian2024large_arxiv2024}\cite{shi2024shield_arxiv2024}\cite{jia2024can_arxiv2024}\cite{li2024fakebench_arxiv2024}\cite{ren2025multimodalreasoningllmswork_arxiv2025}\cite{chen2024textit_arxiv2024}\cite{li2024forgerygpt_arxiv2024}\cite{qin2025towards_arxiv2025}\cite{sun2025towards_arxiv2025}, tier=last, fill=green!20]
                ]
            ]
            [Learning Level, tier=level4, calign=child edge, calign child=4, fill=green!20
                [Advanced Loss Functions, tier=level5, calign=child edge, calign child=1, fill=green!20
                    [\cite{10415851_icdm2023}\cite{10219778_icme2023}\cite{zhang2024domain_acmmcca2024}\cite{farooq2024securing_arxiv2024}, tier=last, fill=green!20]
                ]
                [Reconstruction, tier=level5, calign=child edge, calign child=1, fill=green!20
                    [\cite{khalid2020oc_cvprw}\cite{he2021beyond_ijcai2021}\cite{tian2024real_eccv2024}\cite{cao2022end_cvpr2022}\cite{cai2023marlin_cvpr2023}, tier=last, fill=green!20]
                ]
                [Manipulation Localization, tier=level5, calign=child edge, calign child=1, fill=green!20
                    [\cite{li2020face_cvpr2020}\cite{dang2020detection_cvpr2020}\cite{shuai2023locate_acmmm2023}\cite{dong2023implicitidleakage_cvpr2023}\cite{zhang2023editguard_cvpr2024}\cite{tantaru2024weakly_wacv2024}, tier=last, fill=green!20]
                ]
                [Identity Discrepancy, tier=level5, calign=child edge, calign child=1, fill=green!20
                    [\cite{dong2023implicitidleakage_cvpr2023}\cite{cozzolino2021id_iccv2021}\cite{nirkin2020deepfake_tpami2021}\cite{dong2022protecting_cvpr2022}\cite{huang2023implicit_cvpr2023}\cite{sun2024diffusionfake_nips2024}\cite{hu2024delocate_arxiv2024},tier=last, fill=green!20]
                ]
                [Fine-grained Prediction, tier=level5, calign=child edge, calign child=1, fill=green!20
                    [\cite{guo2023hierarchical_cvpr2023}\cite{guo2024language_ijcv2024}\cite{guarnera2024mastering_acmmcca2024}\cite{guo2023controllable_iccv2023},tier=last, fill=green!20]
                ]
                [Disentangled Representation, tier=level5, calign=child edge, calign child=1, fill=green!20
                    [\cite{liang2022exploring}\cite{yan2023ucf_iccv2023}\cite{lin2024preserving_cvpr2024}\cite{ye_arxiv2024}\cite{yang2023improving_arxiv2023}\cite{ba2024exposing}\cite{zhou2024capture_arxiv2024}, tier=last, fill=green!20]
                ]
                [Knowledge Distillation, tier=level5, calign=child edge, calign child=1, fill=green!20
                    [\cite{woo2022add_aaai2022}\cite{kim2021fretal_cvprw2021}\cite{kim2021cored_mm2021}\cite{lv2023domainforensics_arxiv2023}, tier=last, fill=green!20]
                ]
            ]
        ]
        [Detection for Diffusion Model, tier=level3, calign=child edge, calign child=1, fill=green!20
            [\cite{ricker2022towards_arxiv2022}\cite{Corvi_2023_ICASSP2023}\cite{lorenz2023detecting_iccvw2023}\cite{cheng2024diffusion_arxiv2024}\cite{wang2023dire_iccv2023}\cite{ma2023exposing_icmlw2023}\cite{luo2024lare_cvpr2024}, tier=last, fill=green!20]
        ]
        [Sequential Detection, tier=level3, calign=child edge, calign child=1, fill=green!20
            [\cite{shao2022detecting_eccv2022,shao2023robustseq_arxiv2023,Hong_2024_CVPR}, tier=last, fill=green!20]
        ]
      ]
      [Proactive Detection, tier=level2, fill=green!20
        [Proactive Methods for GANs, tier=level3, calign=child edge, calign child=1, fill=green!20
            [\cite{ruiz2020disrupting_eccvw2020}\cite{yeh2020disrupting_wacvw2020}\cite{huang2021initiative_aaai2021}\cite{wang2022anti_ijcai2022}\cite{aneja2021tafim_eccv2022}\cite{Huang2022_aaai2022}\cite{Asnani2022_cvpr2022}\cite{Asnani2023_cvpr2023}\cite{Wang2023_arxiv2023}\cite{wu2023sepmark_mm2023}\cite{yu2020responsible_iclr2022}\cite{tang2023feature_arxiv2023}\cite{zhang2024dual_tifs2024}\cite{wu2024watermarks_ijcai2024}\cite{qu2024id_arxiv2024}\cite{lan2024facial_arxiv2024}\cite{zhu2024hiding_arxiv2024}\cite{wang2025faceswapguard_arxiv2025}, tier=last, fill=green!20]
        ]
        [Proactive Methods for DMs, tier=level3, calign=child edge, calign child=1, fill=green!20
            [\cite{cui2023diffusionshield_nipsw2024}\cite{fernandez2023stable_iccv2023}\cite{wang2023diagnosis_iclr2024}\cite{zhao2023recipe_arxiv2023}\cite{shim2023leat_arxiv2023}\cite{jeong2024faceshield_arxiv2024}, tier=last, fill=green!20]
        ]
      ]
    ]
    [Multi-modal Detection, tier=level1,fill=gray!10
      [Audio-Visual, tier=level2,fill=blue!20
        [Independent Learning, tier=level3, calign=child edge, calign child=1,fill=blue!20
            [\cite{zhou2021joint}\cite{khalid2021evaluation}\cite{hashmi2022multimodal}\cite{ilyas2023avfakenet}, tier=last,fill=blue!20]
        ]
        [Joint Learning, tier=level3,fill=blue!20
            [Intermediate Fusion, tier=level4, calign=child edge, calign child=1,fill=blue!20
                [Cross-Attention, tier=level5, calign=child edge, calign child=1,fill=blue!20
                    [\cite{zhou2021joint}\cite{10081373}\cite{Marcella_bmvc2024}\cite{yu2023pvass}\cite{oorloff_cvpr2024}\cite{nie2024frade_MM2024}, tier=last,fill=blue!20]
                ]
            ]
            [Late Fusion, tier=level4, calign=child edge, calign child=3,fill=blue!20
                [Concatenation \& \\Addition, tier=level5, calign=child edge, calign child=1,fill=blue!20
                    [\cite{khalid2021evaluation}\cite{hashmi2022multimodal}\cite{kharel2023df}\cite{muppalla2023integrating}\cite{liu2023magnifying}\cite{10034605}\cite{wang2024avt2}, tier=last,fill=blue!20]
                ]
                [Attention, tier=level5, calign=child edge, calign child=1,fill=blue!20
                    [\cite{wang2024avt2}\cite{zou2024cross}\cite{sree2023mis}, tier=last,fill=blue!20]
                ]
                [MLP Mixer Layer, tier=level5, calign=child edge, calign child=1,fill=blue!20
                    [\cite{raza2023multimodaltrace}, tier=last,fill=blue!20]
                ]
                [Multi-task Strategy, tier=level5, calign=child edge, calign child=1,fill=blue!20
                    [\cite{zhou2021joint}\cite{10081373}\cite{muppalla2023integrating}\cite{raza2023multimodaltrace}\cite{zhang2024mfms_MM2024}\cite{yu2024explicit_icme2024},tier=last,fill=blue!20]
                ]
                [Enhanced Representation, tier=level5, calign=child edge, calign child=1,fill=blue!20
                    [\cite{zou2024cross}\cite{sree2023mis}\cite{10081373}\cite{liu2023mcl_tcsvt2023}\cite{liu2023magnifying}, tier=last,fill=blue!20]
                ]
            ]
        ]
        [Matching-based \\Learning, tier=level3, calign=child edge, calign child=1,fill=blue!20
            [\cite{cheng2023voice}\cite{tian2023unsupervised}\cite{Marcella_icip2024}\cite{boldisor2024circumventing_arxiv2024}\cite{feng2023self}, tier=last,fill=blue!20]
        ]
        [Others, tier=level3, calign=child edge, calign child=1,fill=blue!20
            [\cite{cozzolino2023audio}\cite{haq2023multimodal}, tier=last,fill=blue!20]
        ]
      ]
      [Text-Visual, tier=level2, calign=child edge, calign child=1, fill=orange!20
        [\cite{shao2023detecting}\cite{shao2024detecting_tpami2024}\cite{liu2023unified_arxiv2023}\cite{wang2024exploiting}, tier=last, fill=orange!20]
      ]
    ]
  ]
\end{forest}
    \caption{The taxonomy of facial deepfake detection methods.}
    \label{taxonomy}
\end{figure*}

\subsection{Single-modal Detection}
\label{subsec:sing-modal-detection}
In this subsection, we review single-modal detection methods for facial forgery in images and videos, categorizing them into two primary approaches: passive detection and proactive detection.

\subsubsection{Passive Detection Methods}
Passive detection methods identify manipulated videos or images solely by analyzing inherent artifacts or inconsistencies within the content itself, assuming no active intervention or additional markers embedded during content creation. 
Such methods focus on intrinsic features present in the forged media, making them particularly applicable when no auxiliary creation information is available.
In the following subsections, we first examine methods targeting GAN/VAE-generated deepfakes, further categorizing them into basic and generalized detection approaches based on their distinct characteristics. 
We then review recent works addressing detection of diffusion model generated deepfakes and sequential deepfake detection, highlighting emerging trends in these rapidly evolving areas\footnote{Additional details on sequential deepfake detection can be found in the supplementary materials.}.

\paragraph{{Basic Detection} {for GAN/VAE}}  
In facial deepfake detection, artifacts refer to unintended irregularities introduced during AI-driven digital content creation or manipulation. 
Common visual artifacts include lighting and color mismatches~\cite{afchar2018mesonet}, unusual facial textures~\cite{liu2020global_cvpr2020}, irregular frequency patterns~\cite{corvi2023intriguing_cvprw2023}, and inconsistent edge blending~\cite{nguyen2024laa_cvpr2024}. 
Additionally, video-based deepfakes  introduce temporal inconsistencies, arising from frame-to-frame manipulations, and typically manifest as unnatural facial movements~\cite{haliassos2021lips}, identity fluctuations~\cite{liu2023ti2net_wacv2023}, or inconsistent noise patterns~\cite{wang2023noise_aaai2023,chen2024compressed_arxiv2024}. 
Identifying these spatial and temporal artifacts is crucial, as they serve as key indicators that media content is manipulated or generated.


\noindent\textbf{Visual Artifacts-based Methods.}
Early deepfake detection methods fundamentally leveraged visual artifacts, primarily exploiting global texture discrepancies. 
For example, Afchar et al.~\cite{afchar2018mesonet} proposed MesoNet, a streamlined network explicitly designed to detect mesoscopic textural anomalies introduced during forgery. 
Rossler et al.~\cite{rossler2019faceforensics} employed XceptionNet, demonstrating strong capabilities in capturing subtle but widespread artifacts, while Liu et al.~\cite{liu2020global_cvpr2020} further refined global texture analysis using Gram-Net to differentiate genuine facial textures from manipulated ones.
Recognizing limitations inherent in purely global texture analysis, such as reduced sensitivity to localized manipulations and fine-grained details, subsequent research shifted focus to local and attention-based feature extraction strategies. 
Zhao et al.~\cite{zhao2021multi} introduced a multi-attention mechanism specifically designed to emphasize local textural inconsistencies. 
Similarly, Ju et al.~\cite{10246417_tmm2023} developed the Global and Local Feature Fusion framework, strategically selecting image patches to enhance the detection of localized artifacts. 
Nguyen et al.~\cite{nguyen2024laa_cvpr2024} further improved artifact localization accuracy through a multi-task framework incorporating classification, heatmap regression, and self-consistency regularization tasks.

More recently, researchers have actively explored frequency- and filter-based approaches motivated by the hypothesis that generative processes leave distinct spectral signatures that are not easily perceived visually. 
Corvi et al.~\cite{corvi2023intriguing_cvprw2023} identified GAN-specific artifacts in the frequency domain, presenting frequency analysis as an effective alternative to purely spatial features. 
Chen et al.~\cite{chen2024single_arxiv2024} utilized Spatial Rich Model filters to robustly detect subtle manipulation traces at the image patch level. 
Furthermore, Tan et al.~\cite{tan2023rethinking_cvpr2024,tan2024data} specifically targeted artifacts introduced by generative upsampling operations.

While visual artifact-based methods have demonstrated substantial effectiveness, they inherently rely on detectable, stable, and model-specific anomalies, limiting their adaptability to emerging generative techniques. 
Advanced generative models, particularly diffusion-based approaches, continue to minimize visually perceptible artifacts, progressively challenging these detection paradigms. 
Additionally, artifact-based methods often exhibit degraded performance under realistic scenarios, including image compression, resizing, and noise, highlighting their vulnerability to common post-processing operations.
Consequently, future research should explore the development of artifact detection frameworks robust to common image processing operations to enhance real-world applicability.

\noindent\textbf{Consistencies-based Methods.} 
Consistency-based detection methods fundamentally leverage discrepancies within temporal or spatio-temporal patterns across video frames to identify forged content. 
Such methods rest on the critical assumption that generative models, despite advanced capabilities, struggle to maintain fully consistent semantic, spatial, and temporal coherence across synthesized frames. 
The early temporal consistency methods mainly identified basic physiological anomalies, such as unnatural blinking patterns~\cite{8630787} and irregular heartbeat rhythms~\cite{qi2020deeprhythm_mm2020}. 
Subsequently, approaches evolved to capture more complex semantic inconsistencies, notably in lip synchronization~\cite{haliassos2021lips} and latent style dynamics~\cite{choi2024exploiting_cvpr2024}. 

Recent advances further emphasized the nuanced interplay between spatial and temporal dimensions. 
Early transformer-based frameworks, such as the two-stage temporal model by Zheng et al.~\cite{zheng2021exploring_iccv2021}, introduced explicit temporal coherence modeling for improved detection. 
Chu et al.~\cite{chu2025reducedspatialdependencygeneral_icassp2025} further proposed a Spatial Dependency Reduction framework integrating spatial perturbation strategies and temporal transformers to enhance robustness against spatial biases. 
Additional enhancements involved sophisticated disentanglement of spatial and temporal features—Pang et al.~\cite{pang2023mre_tcsvt2023} and Wang et al.~\cite{wang2023altfreezing} explicitly separated these features to heighten sensitivity towards subtle forgery-induced anomalies. 
Chen et al.~\cite{chen2025gc_arxiv2025} employed optical flow residuals combined with global context modeling to precisely identify unnatural spatio-temporal anomalies. 
Additionally, transformer-based architectures incorporating self-supervised and contrastive learning paradigms, such as UniForensics~\cite{fang2024uniforensics_arxiv2024}, FakeSTormer~\cite{nguyen2025vulnerability_arxiv2025}, and CNN-Transformer~\cite{zhangnaco_arxiv2024}, have further demonstrated superior capabilities in capturing fine-grained inconsistencies across both spatial and temporal domains.

In addition, other research efforts have extended consistency analyses to broader domains such as semantic, identity, and noise. 
Semantic inconsistency detection, represented by TALL and TALL++ frameworks~\cite{xu2023tall_iccv2023,xu2024towards}, utilizes graph-based reasoning to pinpoint subtle narrative discrepancies across frames.
Identity consistency detection methods, such as TI2Net~\cite{liu2023ti2net_wacv2023}, explicitly focus on biometric continuity violations, providing robust verification cues. 
Yan et al.~\cite{yan2024generalizing_arxiv2024} further explored subtle inconsistencies related to facial organ positions and shapes, adding nuanced identity-based temporal indicators to detect advanced manipulations. 
Furthermore, latent pattern detection~\cite{latentpattern_acmmma2022}, audio-visual correlation-based self-supervised learning~\cite{haliassos2022leveraging_cvpr2022}, and noise-based methods~\cite{wang2023noise_aaai2023,chen2024compressed_arxiv2024} represent complementary strategies that emphasize fine-grained temporal anomalies and statistical anomalies between authentic and manipulated content.

However, despite significant achievements, consistency-based methods still face inherent theoretical and practical limitations. 
Their effectiveness remains sensitive to variations in video quality, compression artifacts, and subtle temporal perturbations commonly encountered in real-world scenarios. 
Moreover, advanced manipulations incorporating sophisticated temporal coherence strategies can significantly reduce detectable inconsistencies, posing ongoing challenges to robustness. 
To address these issues, future research may explore robust feature extraction mechanisms less affected by video degradation, integrate advanced self-supervised learning paradigms for stronger generalization, and develop unified multi-dimensional frameworks that simultaneously leverage temporal, semantic, and identity-based consistencies for more comprehensive and resilient forgery detection.

\paragraph{{Generalized Detection {for GAN/VAE}}} 
\label{advanced-passive-det-single-modal}
{In the realm of deepfake detection, the generalizability of detection plays an important role in enhancing the adaptability of the detection methods against a wide range of manipulations.
In this subsection, we delve into various categories of approaches, such as input level, model level, and learning level methods.
Each category provides its own perspectives and techniques to tackle the challenge of generalized deepfake detection.}



\noindent\textbf{Input-Level: Data Augmentation}  
Early data augmentation strategies in deepfake detection primarily involved pixel-level perturbations, such as masking or blending regions in genuine images to simulate manipulation artifacts. For example, Wang et al.\cite{Wang2021_cvpr2021} proposed attention-based facial masking, and Zhao et al.\cite{zhao2021learning_iccv2021} introduced an inconsistency-driven image generator. 
These approaches aimed to increase data diversity but were limited in capturing high-level forgery dynamics.
Recent efforts have shifted toward more structured and adaptive augmentation techniques. 
Shiohara et al.\cite{shiohara2022detecting_cvpr2022} and Larue et al.\cite{larue2023seeable_iccv2023} emphasized self-blended or soft-forgery injections to simulate artifact subtleties. 
Li et al.\cite{li2023pixel_tmm2023} enhanced low-quality forgery detection using a bleach generator. 
Cheng et al.\cite{cheng2024can_arxiv2024} and Lin et al.\cite{lin2024faketillmakeit_eccv2024} proposed curriculum-based and bi-level policy-optimized augmentation, respectively, enabling progressive learning of forgery characteristics. 
DiffFake\cite{stamnas2025difffake_wacvw2025} framed augmentation through an anomaly detection lens, generating both global and local forgery cues.

\noindent\textbf{Input-Level: Frequency-space based Methods}  
To address the limitations of RGB-space detection, frequency-domain methods have emerged as a complementary approach for identifying subtle manipulation traces. 
Early works such as Qian et al.\cite{qian2020thinking} and Chen et al.\cite{chen2021local} leveraged frequency statistics and multi-scale similarity to identify anomalies. 
Li et al.~\cite{li2021frequency} proposed single-center loss with adaptive frequency features for enhanced robustness.
More recent studies apply frequency-aware augmentation in both spatial and latent domains. 
Yan et al.\cite{yan2023transcending} and Li et al.\cite{li2024freqblender_arxiv2024} explored latent frequency perturbation and component recombination, respectively, while Song et al.\cite{song2024quality_arxiv2024} introduced quality-centric curriculum learning using frequency-aware transformations.
Subsequent works such as Luo et al.\cite{luo2021generalizing} and Liu et al.~\cite{liu2021spatial} incorporated cross-modal attention and spatial-phase analysis to highlight fine-grained forgery cues.
Advanced approaches increasingly integrate spatial-frequency representations with semantic priors. 
Jeong et al.\cite{jeong2022frepgan}, Tan et al.\cite{tan2024frequency}, and Doloriel et al.\cite{doloriel2024frequency_icassp2024} refined attention and masking strategies in frequency space. 
Li et al.\cite{li2024multiple_arxiv2024} and Baru et al.\cite{baru2024harnessing_arxiv2024} combined multi-kernel and wavelet-based learning, while Peng et al.\cite{peng2025wmamba_arxiv2025} and Yan et al.~\cite{yan2025generalizable_arxiv2025} developed multi-resolution hybrid models integrating DWT, FFT, and ViT backbones.

\ding{42} Recent input-level methods have evolved along two complementary directions: data augmentation and frequency-domain analysis. 
Augmentation strategies have progressed from basic pixel-level perturbations to advanced schemes incorporating latent-space interpolation, curriculum-driven sample synthesis, and bi-level policy optimization, enabling finer control over forgery difficulty and diversity. 
In parallel, frequency-domain approaches have advanced from simple statistical modeling to high-resolution spectrum decomposition, integrating wavelet transforms, spatial-phase learning, and latent-frequency feature engineering. 
These developments have substantially improved model robustness and generalization under cross-dataset and unseen-forgery scenarios.


\noindent\textbf{Model Level: Vision Transformer-Based Approaches} 
Vision Transformers (ViT)~\cite{dosovitskiy2020image_iclr2021} have recently gained substantial traction in deepfake detection, demonstrating strong capabilities for both static image and temporal video analysis~\cite{Wang2024ATS_arxiv2024}. 
Their ability to model global context and long-range dependencies has driven extensive exploration into localized forgery cues, cross-dataset generalization, and efficient adaptation.

\noindent\textit{Spatio-Temporal Feature Extraction.}
Initial transformer-based frameworks emphasized effective spatial-temporal modeling. 
For instance, Saha et al.~\cite{saha2023undercover_iccvw2023} integrated Vision Transformer with Timeseries Transformer to jointly capture manipulated spatial and temporal features. 
Bai et al.~\cite{bai2023aunet_cvpr2023} further introduced an Action Units Relation Transformer to explicitly model facial action unit dynamics, aiding temporal forgery discrimination. 
Zhang et al.~\cite{zhang2023ummaformer_acmmm2023} developed UMMAFormer, employing temporal abnormal attention and cross-attention pyramid modules for precise forgery localization.

\noindent\textit{Enhanced Feature Discrimination.}
Subsequent approaches introduced sophisticated transformer architectures for finer feature discrimination. 
Zhang et al.~\cite{zhang2024genface_arxiv2024} proposed Cross Appearance-Edge Learning to integrate multi-level appearance and edge information, further refining detection granularity. 
Additionally, Fu et al.~\cite{fu2025exploring_aaai2025} explicitly addressed position and content biases inherent in transformers, introducing the UDD framework that employs token-level shuffling and mixing operations within latent transformer spaces, significantly enhancing cross-dataset generalization capabilities.

\noindent\textit{Leveraging Pretrained Transformer Backbones.}
More recently, several works have leveraged pretrained transformer backbones for improved detection robustness.
Pellicer et al.~\cite{pellicer2024pudd_cvprw2024} exploited the pretrained DINOv2 backbone to enhance robustness through prototype-based clustering and matching.
Tsai et al.~\cite{tsai2024understanding_arxiv2024} also utilized pretrained DINOv2 features, demonstrating via frequency-domain analysis that AI-generated facial images exhibit distinct high-frequency artifacts, particularly evident under Gaussian blur perturbations, facilitating training-free deepfake detection.
Chen et al.~\cite{chen2024guided_arxiv2024} proposed the Guided and Fused Frozen CLIP-ViT framework, explicitly guiding CLIP-ViT toward task-specific features through a lightweight Deepfake-Specific Feature Guidance Module and FuseFormer, a Transformer-based fusion mechanism that integrates forgery clues across multiple ViT stages.
Guo et al.~\cite{guo2024face_arxiv2024} leveraged self-supervised pre-training on real-face datasets, introducing a dual-branch fine-tuning framework with a decorrelation constraint, guiding each branch to distinctly capture forgery cues.
Similarly, Li et al.~\cite{li2025deepfakedetectionknowledgeinjection_arxiv2025} introduced Knowledge Injection-based deepfake Detection, preserving real-image knowledge while injecting deepfake-specific understanding via a dedicated knowledge injection module and a multi-task learning strategy to balance real and fake image distributions.

\noindent\textit{Local Attention Mechanisms.}
 {Vision Transformer is known for its inherent tendency to prioritize global features \cite{dosovitskiy2020image_iclr2021,liu2021swin_iccv2021}, recent methods explicitly tackle this by introducing targeted local attention mechanisms to enhance sensitivity to subtle forgery artifacts.}
Nguyen et al.~\cite{nguyen2024fakeformer_arxiv2024} introduced FakeFormer, incorporating a Learning-based Local Attention mechanism to explicitly guide transformers toward local patches prone to subtle forgery artifacts. 
Luo et al.~\cite{luo2024forgery_tcsvt2024} proposed FA-ViT, integrating global, local, and fine-grained adaptive modules to model long-range interactions, expose local inconsistencies, and emphasize genuine face representations via relationship learning. 
Zhang et al. \cite{zhang2025distilled_pr2025} developed a Distilled Transformer Network  featuring a Locally-Enhanced Vision Transformer, significantly enhancing the detection of subtle local artifacts.

\noindent\textit{Parameter-Efficient Transformer Adaptation.}
Recent parameter-efficient tuning and reprogramming strategies {enable efficient} transformer model adaptation in deepfake detection. 
Kong et al.~\cite{Kong2024OpenSetDD_arxiv2024} proposed a forgery style mixture module coupled with lightweight Adapter and LoRA layers integrated into ViTs, diversifying feature statistics while minimizing trainable parameters. 
Shao et al.~\cite{shao2023deepfake_arxiv2023} presented the DeepFake-Adapter, embedding dual-level adapters in ViTs to improve anomaly detection capabilities without altering the backbone. 
Cui et al.~\cite{cui2024forensics_arxiv2024} introduced Forensics Adapter, a lightweight adapter designed to adapt CLIP specifically for face forgery detection by focusing on subtle blending boundaries, achieving superior generalization and significant performance improvements with minimal additional parameters.
Liu et al.~\cite{liu2024mixture_arxiv2024} employed a Mixture-of-Experts approach combined with LoRA modules, achieving enhanced adaptability while preserving pre-trained knowledge. 
Similarly, Kong et al.~\cite{kong2024moe_arxiv2024} introduced MoE-FFD, incorporating both LoRA and Adapters to effectively capture global and local forgery clues. 
Lin et al.~\cite{lin2024standing_aaai2025} developed RepDFD, a visual-language model reprogramming approach using minimal trainable parameters, effectively improving cross-dataset generalization through adaptive visual and text prompts.
Extending the concept of adapter-based transformer tuning for deepfake detection, Guo et al.~\cite{guo2025rethinkingvisionlanguagemodelface_arxiv2025} propose M2F2-Det, which enhances CLIP through forgery prompt learning with hierarchical forged tokens and bridge adapters, effectively integrating CLIP visual features with LLM-generated textual explanations to achieve superior detection accuracy and interpretability.

\noindent\textit{Comprehensive Video-level Detection.}
Additional efforts have broadened Transformer applications to tackle comprehensive video-level forgery detection, including complex non-facial manipulations and motion inconsistencies.
Kundu et al. \cite{kundu2024towards_arxiv2024} proposed a universal synthetic video detector employing Attention-Diversity Loss to analyze entire video frames, effectively identifying various manipulations. 
Furthermore,  Nie et al.\cite{nie2024dip_tmm2024} introduced a video-based transformer framework explicitly analyzing directional inconsistency patterns along horizontal and vertical axes, significantly improving generalization across unseen video forgery methods.

\noindent \textbf{{Model} Level: Vision-Language and Multimodal Large Language Models}
Recent studies have explored the potential of pre-trained Vision-Language Models (VLMs)~\cite{zhang2024vision_tpami2024}, such as CLIP and GPT-4o, highlighting their effectiveness for \textit{universal} detection and enhanced \textit{reasoning} capabilities.

\noindent\textit{Universal Detection:}
Researchers leveraged VLMs to achieve generalized detection across diverse scenarios, gradually evolving from simple feature extraction toward more sophisticated cross-modal alignment and enhanced generalization methods.
Early efforts focused primarily on aligning visual and textual modalities, as demonstrated by Han et al.\cite{han2024towards_arxiv2024}, who combined CLIP’s visual features with a side network to effectively capture spatial-temporal cues. 
Similarly, Lai et al.\cite{Lai2024GMDFGM_arxiv2024} proposed GM-DF, leveraging CLIP-based visual-textual alignment alongside hybrid expert modeling for improved domain generalization.

Subsequent works emphasized leveraging prompt-based learning techniques to further enhance the model’s generalization ability. 
Tan et al.\cite{tan2024c2pclip_arxiv2024} introduced C2P-CLIP, augmenting CLIP with category-specific textual prompts to guide visual feature extraction. 
Lin et al.\cite{lin2024standingshouldersgiantsreprogramming_arxiv2024} advanced this concept further with RepDFD, employing learnable visual perturbations and identity-specific textual prompts, significantly enhancing cross-domain robustness. 
Additionally, Wang et al.~\cite{wang2025knowledge_arxiv2025} implemented knowledge-guided prompt strategies with GPT-4 integration, further improving model adaptability under domain shifts through test-time tuning.

In parallel, a few approaches addressed the challenge of fine-grained forgery detection and localization. 
Kong et al.\cite{kong2023enhancing_wacv2025} presented DeCLIP, combining CLIP’s pretrained features with convolutional decoders to precisely localize manipulated regions within images. 
Extending fine-grained analysis, Zhang et al.\cite{zhang2024mfclipmultimodalfinegrainedclip_arXiv2024} proposed MFCLIP, integrating fine-grained language prompts with multimodal vision encoders, enhancing detailed manipulation detection.

Recently, several studies have underscored the advantage of multimodal semantic understanding in robustness against various perturbations. 
Zhang et al.\cite{zhang2024dpl_accv2024} proposed DPL, utilizing CLIP-based text-image matching to dynamically guide feature extraction under varying image quality conditions. 
Levy et al.\cite{levy2024nearly_arxiv2024} further validated this multimodal advantage, showing that incorporating semantic insights from GPT-4o alongside CLIP significantly mitigates vulnerabilities present in purely visual-based detectors, demonstrating superior adversarial robustness.

\noindent \textit{Reasoning:} 
Advanced Vision-Language Models (VLMs) have significantly improved reasoning capabilities and interpretability in deepfake detection.
Tsigos et al.\cite{tsigos2024towards_icmrw2024} introduced a framework to evaluate visual explanation accuracy under adversarial perturbations, while Sun et al.\cite{sun2023towards_arxiv2023} enriched textual and visual training data to strengthen model interpretability.
Zhang et al.\cite{zhang2024common_arxiv2024} reformulated deepfake detection as a visual question-answering task using a BLIP-based architecture \cite{li2022blip_icml2022}, allowing models to explain their reasoning explicitly.
Huang et al.\cite{huang2024ffaa_arxiv2024} advanced this concept by incorporating GPT-4o for sophisticated reasoning and robust multi-answer decision-making processes.
Most recently, Lian et al.~\cite{lian2024large_arxiv2024} proposed ForgeryTalker, a specialized vision-language framework extending InstructBLIP \cite{dai2023instructblip_nips2023}, which simultaneously localizes manipulated facial regions and generates detailed textual explanations through targeted forgery-aware prompting mechanisms.

More recently, researchers have leveraged Multimodal Large Language Models to significantly strengthen both detection accuracy and interpretability. 
Shi et al.\cite{shi2024shield_arxiv2024}, Shan et al.\cite{jia2024can_arxiv2024}, Li et al.\cite{li2024fakebench_arxiv2024}, and Ren et al. \cite{ren2025multimodalreasoningllmswork_arxiv2025} provided comprehensive validations, confirming the effectiveness of MLLMs in deepfake detection. 
Chen et al.~\cite{chen2024textit_arxiv2024} developed X²-DFD,  integrating external detectors to significantly boost accuracy and explanation quality. 
Li et al.~\cite{li2024forgerygpt_arxiv2024} introduced ForgeryGPT, combining MLLMs and mask-aware extractors for detailed artifact localization and explanations. 
Qin et al.~\cite{qin2025towards_arxiv2025} proposed DFA-GPT, enabling interactive detection, explanations, and conversational analyses through instruction-tuned MLLMs. 
Sun et al.~\cite{sun2025towards_arxiv2025} further advanced interpretability through structured prompts that generate precise textual annotations.
He et al.~\cite{he2025vlforgeryfacetriaddetection_arxiv2025} recently introduced {VLForgery}, a unified multimodal forensic framework that leverages MLLMs to simultaneously perform detection, localization, and attribution of AI-generated faces. 
Specifically, {VLForgery} employs an Extrinsic knowledge-guided Chain-of-thought (EkCot) mechanism, which systematically integrates low-level visual comparisons with external knowledge about generative processes, significantly enhancing forensic interpretability and analytical precision.

Recognizing the substantial computational overhead associated with fine-tuning large models, recent works have explored lightweight, prompt-based tuning strategies. 
Chakraborty et al.~\cite{chakraborty2025truthlensatrainingfreeparadigmdeepfake_arxiv2025} proposed TruthLens, a training-free paradigm that leverages advanced large vision-language models such as ChatUniVi \cite{jin2024chat_cvpr2024}, LLaVA 1.5 \cite{liu2023visual_nips2023}, BLIP-2 \cite{li2023blip_icml2023}, and CogVLM \cite{wang2024cogvlm_nips2024} for deepfake classification and interpretable reasoning without explicit training on synthetic datasets. 
Yu et al.~\cite{yu2025unlockingcapabilitiesvisionlanguagemodels_arxiv2025} further advanced this direction, proposing a generalizable and explainable deepfake detection framework that utilizes large vision-language models enhanced with knowledge-guided forgery prompts and lightweight prompt tuning, achieving superior detection and localization capabilities with minimal tuning overhead. 
These approaches represent a significant shift towards flexible, scalable, and interpretable solutions, aligning well with practical deployment requirements.

\ding{42} Recent advancements in Vision Transformers (ViTs) have significantly improved the modeling of spatial, temporal, and structural inconsistencies in facial forgery detection. 
However, their inherent bias toward global features often leads to insufficient sensitivity to subtle local artifacts, and their robustness degrades under domain shifts or fine-grained perturbations. 
Parameter-efficient tuning strategies, such as adapters and LoRA, provide practical scalability but may underperform in open-set or data-scarce scenarios without task-specific adaptation.
Vision-Language Models  and Multimodal Large Language Models have introduced new paradigms for interpretable and cross-modal deepfake detection. 
While these models exhibit strong generalization and semantic reasoning capabilities, aligning textual priors with visually grounded evidence remains challenging—particularly under adversarial perturbations or distributional mismatches. 
Moreover, the heavy reliance on pretrained embeddings not designed for forensic tasks limits their reliability in high-stakes applications.

Looking forward, future research should focus on hybrid architectures that integrate localized spatial-frequency cues with multimodal semantic representations. 
There is also a pressing need for standardized evaluation protocols that go beyond accuracy to assess robustness, adaptability, and interpretability across diverse manipulation types and modalities.



\noindent\textbf{Learning Level: Advanced Loss Functions} 
Traditional deepfake detection approaches predominantly rely on Softmax loss, yet recent studies highlight its limitations in effectively distinguishing subtle forgery artifacts. 
Yin et al.\cite{10415851_icdm2023} proposed a dual-domain margin loss that decouples embedding vectors into magnitude and directional components, applying distinct margin penalties within both angular and Euclidean spaces to significantly enhance inter-class separability. 
Zhang et al.\cite{10219778_icme2023} further improved embedding discrimination by introducing a fractional center loss, explicitly tightening intra-class compactness for real images. 
Extending their previous work, Zhang et al.\cite{zhang2024domain_acmmcca2024} designed a dual-level center loss incorporating both patch-level and image-level constraints, simultaneously refining local and global embedding distributions to robustly differentiate real from fake images. 
Complementarily, Farooq et al.\cite{farooq2024securing_arxiv2024} developed DBaGNet, utilizing a triplet margin loss to jointly optimize deep identity, behavioral, and geometric descriptors, systematically capturing complex multi-modal inconsistencies and achieving improved generalization across diverse deepfake datasets.

\noindent\textbf{{Learning Level: }Reconstruction}
Reconstruction-based methods exploit inherent discrepancies between original and reconstructed representations to identify forged content, evolving progressively from basic anomaly detection toward sophisticated frameworks capable of capturing subtle, generalized forgery-specific patterns.
Early studies primarily relied on detecting reconstruction errors obtained by models trained solely on authentic data. 
For instance, Khalid et al.\cite{khalid2020oc_cvprw} introduced a One-class Variational Autoencoder, which models genuine facial image distributions exclusively, making forged images discernible through higher reconstruction anomalies. 
Expanding upon this, He et al.\cite{he2021beyond_ijcai2021} employed sophisticated re-synthesis methods such as super-resolution to isolate residual artifacts, thus extracting more discriminative forgery-specific signals.
Recently, Tian et al.~\cite{tian2024real_eccv2024} introduced Real Appearance Modeling, which further refines reconstruction methods by explicitly modeling real facial appearance from faces subjected to subtle perturbations, effectively capturing subtle residual artifacts and significantly enhancing generalization to unseen forgeries.

Subsequently, researchers integrated advanced constraints and metric-based strategies to enhance reconstruction accuracy and discriminative power. 
Cao et al.\cite{cao2022end_cvpr2022} proposed an end-to-end framework that combines reconstruction with explicit metric constraints, significantly reinforcing the model’s capacity to distinguish authentic from manipulated images by better aligning feature representations. 
More recently, Cai et al. introduced MARLIN~\cite{cai2023marlin_cvpr2023}, a self-supervised masked autoencoder framework leveraging extensive unlabeled online video data to reconstruct spatio-temporal details from densely masked facial regions. 
By learning robust facial feature representations through this extensive reconstruction process, MARLIN effectively captures generalized manipulation artifacts, significantly improving robustness and generalization performance across diverse deepfake detection scenarios.

\noindent\textbf{{Learning Level: }Manipulation Localization} 
Manipulation localization aims to precisely identify regions altered in deepfake samples, providing insights into specific modifications and potential intent behind manipulations~\cite{Asnani2023_cvpr2023}. 
Early approaches, such as Face X-ray~\cite{li2020face_cvpr2020}, utilized subtle blending boundaries indicative of manipulation. 
Dang et al.\cite{dang2020detection_cvpr2020} improved accuracy by integrating attention mechanisms for refined feature extraction, while Shuai et al.\cite{shuai2023locate_acmmm2023} combined multimodal noise residuals and multi-scale features for enhanced forgery region localization. 
Addressing implicit identity leakage, Dong et al.\cite{dong2023implicitidleakage_cvpr2023} emphasized authenticity verification alongside localization, significantly enhancing model generalization. 
Additionally, proactive embedding techniques such as EditGuard\cite{zhang2023editguard_cvpr2024} introduced imperceptible watermarking for accurate localization. 
More recently, Tantaru et al.\cite{tantaru2024weakly_wacv2024} proposed a weakly supervised method, improving scalability and localization accuracy without extensive pixel-level annotations.
These  advancements underscore the evolution from explicit boundary-based approaches toward sophisticated, annotation-efficient, and identity-aware localization techniques

\noindent \textbf{{Learning Level: }Fine-grained Prediction} 
Fine-grained deepfake detection methods have emerged as a critical strategy to address the increasing complexity and diversity of deepfake techniques, enabling detectors to perform detailed hierarchical classification beyond binary real-or-fake predictions.
Guo et al.\cite{guo2023hierarchical_cvpr2023,guo2024language_ijcv2024} proposed a hierarchical detection framework that classifies forgeries progressively, from broad categories down to specific manipulation techniques, significantly enhancing the interpretability and granularity of detection results. 
Similarly, Guarnera et al.\cite{guarnera2024mastering_acmmcca2024} developed a multi-level classification method, first distinguishing real images from synthetic ones, then differentiating GAN-generated images from Diffusion Model-generated ones, and finally pinpointing the specific generative architectures used. 
Additionally, Guo et al.~\cite{guo2023controllable_iccv2023} introduced a controllable guide-space strategy that finely partitions forgery domains to improve feature discrimination, effectively enhancing model generalization across diverse forgery scenarios.

\noindent\textbf{{Learning Level: }Identity Discrepancy} 
Identity discrepancy methods shift deepfake detection from general artifact recognition toward identifying biometric inconsistencies in manipulated identities. 
Early work by Cozzolino et al.\cite{cozzolino2021id_iccv2021} combined facial feature extraction, temporal analysis, and generative adversarial networks to highlight biometric anomalies. 
Nirkin et al.\cite{nirkin2020deepfake_tpami2021} detected discrepancies by comparing identity embeddings between inner and outer facial regions, a strategy further enhanced by Dong et al.\cite{dong2022protecting_cvpr2022} through an Identity Consistency Transformer employing consistency losses. 
Huang et al.\cite{huang2023implicit_cvpr2023} leveraged implicit identity features via novel contrastive loss functions to sharpen intra- and inter-class distinctions. 
More recently, DiffusionFake~\cite{sun2024diffusionfake_nips2024} utilized a pretrained Stable Diffusion model to explicitly disentangle source-target identity features, significantly enhancing generalization to unseen forgeries. 
Hu et al.~\cite{hu2024delocate_arxiv2024} introduced \textit{Delocate}, a two-stage detection framework that exploits identity consistency to localize manipulated facial regions, thereby improving robustness against forgeries with randomly placed manipulations.

\noindent{\textbf{Learning Level: Disentangled Representation} 
Disentanglement learning addresses the issue that deepfake detectors may inadvertently learn representations containing both forgery-related and content-related information. Such mixed representations often lead detectors to overfit by relying on irrelevant cues like face identity or background characteristics~\cite{liang2022exploring, yan2023ucf_iccv2023}. 
To mitigate this, researchers have explored methods to explicitly decouple features into task-relevant (forgery-specific) and task-irrelevant (content or identity) groups, ensuring that detection decisions are based exclusively on forgery artifacts.

Early disentanglement approaches primarily leveraged multi-task and contrastive learning strategies to achieve semantic-level decoupling. 
Liang et al.\cite{liang2022exploring} first proposed a dual-encoder framework that explicitly separates forgery and content-related embeddings, using contrastive constraints to ensure effective semantic disentanglement. 
Yan et al.\cite{yan2023ucf_iccv2023} improved upon this by proposing the Universal Content-Forgery framework, which provided stronger constraints to distinctly isolate forgery-specific features, thereby significantly enhancing model generalization across various forgery types. 
Lin et al.\cite{lin2024preserving_cvpr2024} further expanded disentanglement by explicitly addressing demographic biases, decoupling demographic attributes from forgery artifacts, thus improving fairness alongside detection robustness. 
Similarly, Ye et al.\cite{ye_arxiv2024} introduced an adaptive high-pass filtering module combined with a two-stage training scheme to refine the isolation of forgery semantics, effectively reducing overfitting and boosting generalizability.

Subsequent research moved toward finer-grained approaches using attention mechanisms and information-theoretic frameworks. 
Yang et al.\cite{yang2023improving_arxiv2023} conceptualized different forgery techniques as separate domains and developed the Deep Information Decomposition framework. 
It employs adaptive attention modules to selectively decompose facial information into forgery-specific categories, resulting in more robust detection across various forgery datasets.
Differing from prior semantic-based methods, Ba et al.\cite{ba2024exposing} introduced a local-level disentanglement strategy. 
Their framework leverages a dedicated disentanglement module guided by Local Information Loss, extracting multiple non-overlapping local forgery features. 
These features are subsequently combined, guided by a Global Information Loss, to further eliminate task-irrelevant information.
Most recently, Zhou et al.~\cite{zhou2024capture_arxiv2024} introduced a progressive disentanglement framework that systematically disentangles identity and artifact features through coarse-to-fine strategies. 
They utilized Information Bottleneck theory \cite{10438074_tpami2024} to explicitly compress correlations between identity and artifact representations, significantly improving detection generalization and robustness.

\noindent \textbf{{Learning Level: }Knowledge Distillation}
Knowledge distillation~\cite{gou2021knowledge_ijcv2021} techniques enhance deepfake detection by effectively transferring generalizable representations and decision-level knowledge from powerful teacher models to lightweight or adaptive student models. 
Woo et al.\cite{woo2022add_aaai2022} addressed compressed forgery detection by utilizing frequency attention distillation to recover high-frequency information and correlation distillation via Sliced Wasserstein Distance, significantly improving performance on low-quality forgeries. 
Addressing limited source data availability, Kim et al.\cite{kim2021fretal_cvprw2021} introduced Feature Representation Transfer Adaptation Learning, which efficiently adapts models to new forgery types through latent feature and logit-level distillation. 
Extending this, their CoReD framework~\cite{kim2021cored_mm2021} employed knowledge distillation to mitigate catastrophic forgetting in sequential domain adaptation scenarios, ensuring stable performance across changing forgery domains. 
More recently, Lv et al.~\cite{lv2023domainforensics_arxiv2023} proposed a bidirectional adaptation strategy involving supervised distillation and adversarial training for forward adaptation, complemented by self-distillation for backward adaptation, effectively enhancing cross-domain generalization even without labeled  data.

\ding{42} Learning-level strategies in deepfake detection have evolved significantly, advancing from conventional classification loss functions to highly specialized techniques that promote discriminative, generalizable, and interpretable representations.
However, despite these advances, several open challenges persist. 
For example, loss function enhancements, such as margin-based and center-aware objectives, often rely on well-separated distributions, which may not hold under real-world domain shifts. 
Disentanglement frameworks offer conceptual clarity by decoupling task-relevant and irrelevant features, yet their performance is sensitive to architectural assumptions and often requires strong supervision or auxiliary labels.
Reconstruction-based approaches and identity discrepancy methods provide powerful avenues for modeling anomaly-like residuals and biometric inconsistencies. 
However, their robustness is contingent on precise facial alignment and may degrade in unconstrained scenarios involving occlusions or cross-modal perturbations. 
Localization and fine-grained prediction techniques enhance interpretability, but often at the cost of annotation complexity and increased model capacity. 
Knowledge distillation frameworks improve scalability and deployment efficiency, yet remain bottlenecked by teacher quality, domain alignment, and the difficulty of preserving detection sensitivity through compression.

Moving forward, a promising direction lies in integrating these learning paradigms into unified, task-adaptive frameworks that dynamically balance discriminability, generalization, and interpretability. 
Moreover, future efforts should address the lack of standardized benchmarks for evaluating cross-task consistency, robustness to real-world manipulations, and the trade-offs between model size, inference cost, and forensic reliability.
\paragraph{{Detection for Diffusion Model}}
Diffusion models (DMs) have rapidly emerged as state-of-the-art generative tools due to their iterative denoising mechanisms.
However, their increased fidelity and diversity introduce substantial challenges for forgery detection, as traditional detectors, originally designed for GAN-based manipulations, struggle to generalize.

Early efforts focused on identifying distinct statistical traces introduced by diffusion processes. 
Ricker et al.~\cite{ricker2022towards_arxiv2022} demonstrated that retraining existing GAN-based detectors on DM-generated samples significantly improved detection performance and enabled cross-model generalization. 
Similarly, Corvi et al.~\cite{Corvi_2023_ICASSP2023} discovered frequency-domain artifacts, such as unique spectral peaks specific to models like GLIDE~\cite{nichol2021glide_arxiv2021} and Stable Diffusion~\cite{rombach2022high}, validating the efficacy of frequency-space analysis.

Moving beyond spectral signatures, subsequent research introduced intrinsic feature-based methods. 
Lorenz et al.\cite{lorenz2023detecting_iccvw2023} employed lightweight multi-local intrinsic dimensionality measures, originally designed for adversarial detection, effectively identifying synthetic images and tracing their diffusion model sources. 
Cheng et al.\cite{cheng2024diffusion_arxiv2024} proposed Edge Graph Regularization, integrating edge structures derived via Sobel operators \cite{zhao2008sobel_icip2008} into the model training, simultaneously analyzing image features and structural cues.

Complementing these efforts, reconstruction-based approaches exploit the generative characteristics of diffusion models.
Wang et al.\cite{wang2023dire_iccv2023} introduced Diffusion Reconstruction Error (DIRE), showing that DM-generated images can be reconstructed more faithfully by diffusion models compared to real images. 
Extending this idea, Ma et al.\cite{ma2023exposing_icmlw2023} developed SeDID to address DIRE’s limitations, leveraging intermediate diffusion steps to better capture discriminative features. 
By performing detection directly in the latent space, Luo et al.~\cite{luo2024lare_cvpr2024} effectively simplified the reconstruction approach, significantly lowering computational overhead and improving detection accuracy.

\ding{42} Despite these advances, robust and generalizable detection of diffusion-generated content remains an open challenge, particularly as DMs evolve toward controllable, high-resolution, and multi-modal generation. 
Future work should explore hybrid detection paradigms that combine statistical priors, generative inconsistencies, and semantic reasoning, along with benchmarks that reflect real-world deployment scenarios.

\subsubsection{Proactive Detection Methods}
Proactive defense methods aim to protect content against deepfake manipulation by embedding subtle adversarial signals into original images or videos, {consequently disabling malicious manipulation on pristine media}. 
Unlike passive methods that analyze content after forgery occurs, proactive approaches strategically insert perturbations that remain visually imperceptible yet effectively disrupt the functionality of deepfake generation models or facilitate provenance tracing post-manipulation. 
Figure \ref{passive-proactive}  illustrates the key conceptual differences between proactive and passive detection paradigms. 
In the following subsections, we separately explore recent proactive defense strategies specifically tailored for GAN-based and DM-based deepfakes.

\begin{figure*}
    \centering
    \includegraphics[width=\textwidth]{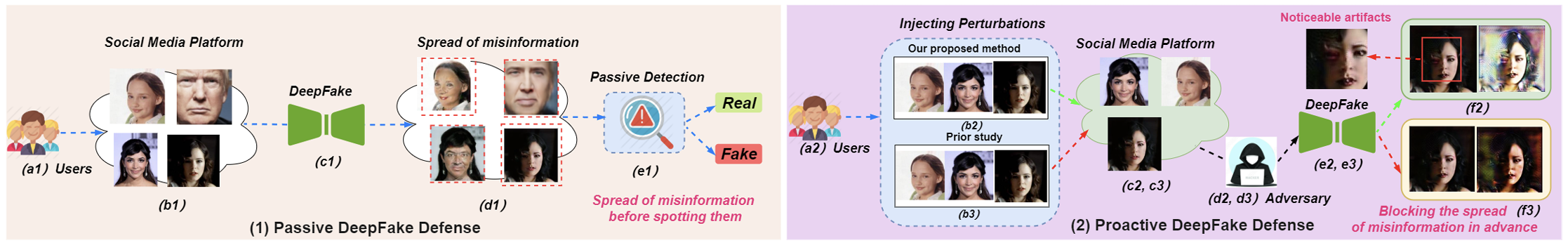}
    \caption{The comparison between passive and proactive deepfake detection methods. Figure from \cite{wang2022anti_ijcai2022}.}
    \label{passive-proactive}
\end{figure*}

\paragraph{Proactive Methods for GANs}
Proactive defense methods for GAN-based deepfake generation primarily focus on embedding robust perturbations to disrupt unauthorized facial manipulations and incorporating traceability mechanisms to track provenance.  
Early proactive defenses mainly utilized generic adversarial strategies designed to induce significant visual degradation in deepfake outputs. 
Ruiz et al.\cite{ruiz2020disrupting_eccvw2020} introduced versatile adversarial attacks combined with adversarial training to explicitly degrade visual quality of manipulated faces. 
Yeh et al.\cite{yeh2020disrupting_wacvw2020} further refined this strategy through customized adversarial loss functions, making unauthorized manipulations visibly distorted and easily detectable.

Subsequent works significantly improved the robustness and specificity of adversarial perturbations. 
Huang et al.\cite{huang2021initiative_aaai2021} employed alternating training with poison perturbation generators, enhancing the stability of injected perturbations. 
Wang et al.\cite{wang2022anti_ijcai2022} specifically utilized Lab color space manipulations to create visually robust adversarial patterns.
Aneja et al.\cite{aneja2021tafim_eccv2022} proposed tailored perturbations designed to redirect manipulation results toward predefined outcomes, achieving finer-grained defensive control.
Huang et al.\cite{Huang2022_aaai2022} developed Cross-Model Universal Adversarial Watermarks to provide universal applicability across diverse GAN architectures, while Asnani et al.\cite{Asnani2022_cvpr2022,Asnani2023_cvpr2023} leveraged real-image templates for enhanced  accuracy.

Recognizing the increasing importance of traceability, recent studies have integrated provenance-tracking capabilities within proactive methods. 
Wang et al.\cite{Wang2023_arxiv2023} and Wu et al.\cite{wu2023sepmark_mm2023} embedded robust and traceable perturbations directly into original images. 
Yu et al.\cite{yu2020responsible_iclr2022} uniquely embedded identifiable fingerprints within generative models themselves, enhancing forensic analysis. Tang et al.\cite{tang2023feature_arxiv2023} introduced universal latent-space disruptors to strengthen cross-model resilience. 
Zhang et al.\cite{zhang2024dual_tifs2024} designed Dual Defense, employing invisible watermarks capable of simultaneously disrupting manipulation processes and ensuring robust provenance tracking. 
Wu et al.\cite{wu2024watermarks_ijcai2024} further developed AdvMark, effectively resolving conflicts between watermark robustness and forgery detection via a plug-and-play framework.

Most recently, proactive defenses have explicitly targeted critical initial steps in GAN-based deepfake pipelines, such as face detection and identity encoding, to prevent forgeries at their earliest stages. 
Qu et al.\cite{qu2024id_arxiv2024} specifically destroyed identity semantics within facial images. 
Lan et al.\cite{lan2024facial_arxiv2024} introduced dynamic, facial-feature-based watermarks enhancing robustness against sophisticated forgery attempts. 
Similarly, FacePoison~\cite{zhu2024hiding_arxiv2024} and FaceSwapGuard~\cite{wang2025faceswapguard_arxiv2025} strategically interfered with critical preprocessing steps, effectively neutralizing deepfake generation before it even begins.

\paragraph{Proactive Methods for DMs}
Distinct from GANs, proactive defense strategies for diffusion models explicitly leverage their unique iterative denoising and conditional attention mechanisms, presenting fundamentally different design considerations.
Early diffusion-specific defenses emphasized robust watermarking within the latent diffusion process, embedding uniform and traceable signals directly into latent spaces or diffusion steps~\cite{cui2023diffusionshield_nipsw2024,fernandez2023stable_iccv2023,wang2023diagnosis_iclr2024,zhao2023recipe_arxiv2023}.
Recent strategies further strengthened the robustness and generalization of watermarking by strategically disrupting latent encodings~\cite{shim2023leat_arxiv2023}, effectively hindering cross-model diffusion manipulations.
Specifically addressing facial forgery, the latest DM proactive defenses explicitly target critical conditional attention mechanisms unique to DMs. 
For example, FaceShield~\cite{jeong2024faceshield_arxiv2024} injects minimal adversarial noise specifically into conditional attention and feature extraction modules, maintaining human imperceptibility and operational robustness.
These diffusion-specific proactive strategies thus distinctly focus on targeted latent-space perturbations and conditional mechanisms, effectively countering the specialized generative processes inherent in diffusion models.

\subsubsection{Discussion: Passive and Proactive Detection}
Deepfake detection research has significantly evolved, transitioning from initial passive methods relying on straightforward visual and temporal artifacts to sophisticated proactive defenses, highlighting essential theoretical insights and practical implications.

Early passive detection approaches primarily exploited identifiable visual anomalies and temporal inconsistencies. 
These methods progressively integrated advanced frequency domain analyses, semantic reasoning via Vision Transformers, and enhanced interpretability through multimodal large language models. 
However, passive methods inherently rely on reactive assumptions: that subtle anomalies will always remain detectable, an assumption increasingly challenged by rapid advancements in generative technologies, particularly diffusion models. 
This inherent limitation is further exacerbated under realistic conditions that involve compression, subtle manipulation, or adversarial post-processing.
To overcome these fundamental constraints, future research should consider adaptive approaches such as continual learning and meta-learning, enabling models to dynamically adapt to evolving forgery strategies and unseen generative architectures, thereby significantly enhancing long-term generalization.

In contrast, proactive detection has advanced from generic visual degradation strategies toward targeted interventions that disrupt foundational generative steps, such as identity encoding and conditional attention mechanisms in diffusion models. 
However, proactive defenses continue to face significant practical limitations: (1) adversarial perturbations may inadvertently degrade perceptual quality, diminishing user acceptance; (2) embedded proactive signals, such as watermarks or perturbations, risk exposing sensitive user attributes, raising critical privacy concerns; and (3) ensuring broad applicability across diverse generative architectures remains challenging due to model-specific vulnerabilities.
Given these complementary strengths and weaknesses, future research would greatly benefit from rigorously exploring integrated passive-proactive frameworks. 
Specifically, constructing comprehensive datasets that systematically incorporate proactively perturbed samples with natural manipulations could provide robust and diversified training signals. 
Moreover, designing principled modeling techniques, such as joint optimization via multitask learning of passive artifacts and proactive patterns, could enhance model robustness and adaptability. 
Lastly, deploying forensic solutions via hierarchical pipelines, where proactive signals enable rapid initial detection followed by detailed passive analyses, offers practical advantages in balancing computational efficiency and forensic thoroughness.

\subsection{Multi-modal Deepfake Detection} 
With advancements in deepfake technologies, manipulated content has evolved from single-modal (e.g., visual-only) to multi-modal forms, where simultaneous falsification across visual, auditory, or textual modalities occurs, resulting in increasingly convincing forgeries.
Compared to single-modal manipulations, multi-modal deepfakes pose distinct challenges due to their {intrinsically heterogeneous input, along with }enhanced semantic coherence and cross-modal consistency, complicating traditional detection methods.
In this subsection, we review recent detection methods designed specifically for complex multi-modal scenarios, addressing simultaneous manipulations across visual-audio and visual-text modalities.

\subsubsection{Audio-Visual Deepfake Detection}
Audio-visual deepfakes involve the manipulation of either or both the audio and visual streams within a video. 
The concurrent alteration of these two modalities significantly heightens the realism and believability of the manipulated content, potentially exacerbating ethical and societal risks. 
For instance, sophisticated audio-visual deepfakes may convincingly depict prominent figures, such as politicians, delivering entirely fabricated statements with realistic lip synchronization, thus misleading viewers and spreading misinformation effectively.
However, detecting such multi-modal manipulations poses unique challenges due to the heterogeneous nature and inherent modality gap between audio and visual signals. Addressing these complexities requires methods capable of effectively leveraging complementary and cross-modal relationships. 
Recent research commonly adopts two-stream architectures to exploit audio and visual inputs separately or jointly.
As illustrated in Figure \ref{fig:similarity}, existing methodologies for audio-visual deepfake detection can be broadly categorized into three paradigms: \textit{independent learning}, \textit{joint learning}, and \textit{matching-based learning}. Each of these paradigms adopts distinct strategies for integrating audio-visual information, thereby enabling robust detection of sophisticated, concurrent manipulations across these modalities.

\begin{figure*}[!htbp]
  \centering
  \begin{subfigure}{0.34\textwidth}
      \includegraphics[width=\textwidth]{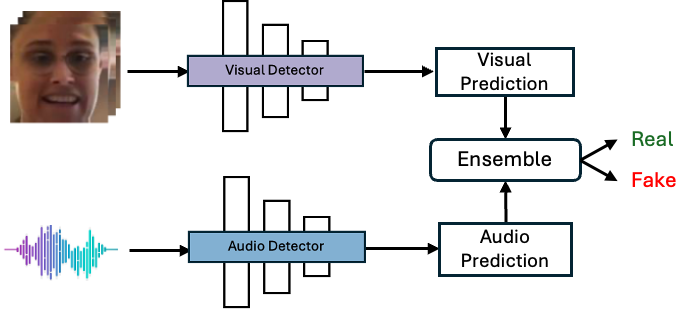}
      \caption{Independent learning for audio-visual deepfake detection.}
      \label{av-indpt}
  \end{subfigure}
  \hfill
    \begin{subfigure}{0.27\textwidth}
      \includegraphics[width=\textwidth]{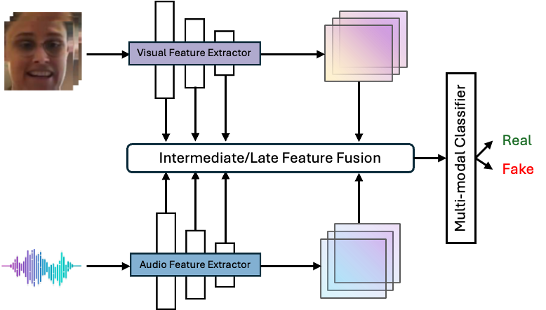}
      \caption{Joint learning for audio-visual deepfake detection.}
      \label{av-joint}
  \end{subfigure}
  \hfill
    \begin{subfigure}{0.32\textwidth}
      \includegraphics[width=\textwidth]{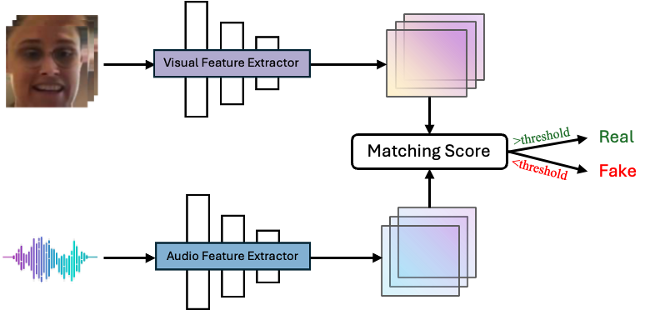}
      \caption{Matching-based learning for audio-visual deepfake detection.}
      \label{av-match}
  \end{subfigure}
  \caption{Three main paradigms for multi-modal audio-visual deepfake detection.
  }
  \label{fig:similarity}
\end{figure*}

\noindent \textbf{Independent Learning} 
The independent learning paradigm processes audio and visual modalities separately, each generating independent predictions, which are subsequently fused at the decision-level to determine content authenticity~\cite{zhou2021joint,khalid2021evaluation,hashmi2022multimodal,ilyas2023avfakenet}.
Zhou et al.\cite{zhou2021joint} proposed a straightforward baseline using two independent convolutional neural networks for audio and visual modality analysis, with results combined at the decision level. 
In an early evaluation on the FakeAVCeleb benchmark\cite{khalid2021fakeavceleb}, Khalid et al.\cite{khalid2021evaluation} adopted an ensemble strategy by aggregating predictions from individually trained unimodal models. 
Extending this concept, Hashmi et al.\cite{hashmi2022multimodal} designed separate audio, visual, and audio-visual networks, integrating their predictions through ensemble learning. 
Similarly, AVFakeNet~\cite{ilyas2023avfakenet} utilized Swin-Transformer~\cite{liu2021swin_iccv2021} architectures for independent audio and visual feature extraction, integrating outputs for final multimodal classification.
Despite its intuitive simplicity, independent learning may neglect intrinsic cross-modal correlations, potentially limiting the detection performance by failing to capture subtle yet crucial relationships between audio and visual modalities.

\noindent\textbf{Joint Learning} 
The joint learning paradigm explicitly integrates audio and visual modalities to effectively exploit their inter-modal relationships, enhancing the robustness of deepfake detection. 
Typically, a joint learning network involves two separate modality-specific branches for extracting audio and visual features, and a dedicated fusion module to integrate these features at different stages. 
Depending on the fusion stage, joint learning approaches can be further categorized into \textit{intermediate fusion} and \textit{late fusion}. 
Additionally, a few methods utilize \textit{multi-task learning} or \textit{advanced representation} to cross-modal representations.

\noindent \textit{Intermediate Fusion:} 
Unlike independent learning methods that integrate predictions derived from separately extracted modality-specific features, intermediate fusion methods emphasize earlier-stage interactions between modalities, under the hypothesis that merely combining high-level abstract features is insufficient to capture rich inter-modal relationships. 
Zhou et al.\cite{zhou2021joint} first introduced intermediate fusion in audio-visual deepfake detection, handling scenarios where either or both modalities might be manipulated. 
They proposed a sync-stream network, where intermediate modality-specific features interact via a cross-attention mechanism to mine deeper cross-modal correlations. 
Similarly, Yang et al.\cite{10081373} utilized a Transformer architecture, where modality-specific features extracted from temporal-spatial encoders are integrated via a decoder equipped with bidirectional cross-attention.

Marcella et al.~\cite{Marcella_bmvc2024} explicitly computed {audio-}spatial attention maps to mitigate interference from irrelevant visual regions (e.g., hair, background), allowing the intermediate audio and visual features to attend to each other at multiple intermediate stages, thereby enhancing fusion effectiveness.
Yu et al.\cite{yu2023pvass} further enriched this intermediate interaction by generating augmented visual views for better alignment with audio information, achieving robust audio-visual representations. 
Oorloff et al.\cite{oorloff_cvpr2024} leveraged a self-supervised two-stage framework, employing intermediate interactions between visual and audio streams via tailored latent modules, explicitly enhancing multi-modal correspondence. 
Most recently, Nie et al.\cite{nie2024frade_MM2024} explicitly addressed inherent semantic gaps between modalities through an audio-distilled cross-modal interaction module in each Transformer block, further bridging the audio-visual domain gap during intermediate feature extraction.
Collectively, these intermediate fusion methods consistently promote rich, semantically coherent multi-modal interactions, significantly improving the accuracy and robustness of audio-visual deepfake detection.

\noindent \textit{Late Fusion:} 
Distinct from independent learning, which integrates predictions at the decision-making stage, and intermediate fusion, emphasizing mid-level cross-modal interactions, late fusion methods specifically integrate modality-specific features at the final stage before classification. 
Common late fusion techniques include simple concatenation or averaging of features \cite{khalid2021evaluation}\cite{hashmi2022multimodal}\cite{kharel2023df}\cite{muppalla2023integrating}\cite{liu2023magnifying}\cite{10034605}\cite{wang2024avt2}, as well as more advanced methods such as attention mechanisms \cite{zou2024cross}\cite{sree2023mis} and MLP mixers \cite{raza2023multimodaltrace}.

Early approaches typically used concatenation for integrating audio-visual features. 
Khalid et al.\cite{khalid2021evaluation} adapted single-modal models from existing multi-modal tasks by concatenating modality-specific features for multi-modal deepfake detection. 
Similarly, Hashmi et al.\cite{hashmi2022multimodal} employed parallel CNN-based audio and visual branches and concatenated extracted features at the late stage. 
DF-TransFusion~\cite{kharel2023df} further refined this paradigm by employing a self-attention branch for facial features and a cross-attention branch for audio-lip synchronization, with their outputs concatenated for final classification.

Beyond concatenation, dynamic weighted fusion has emerged as an effective strategy. 
Cai et al.\cite{10034605} used dynamically weighted sums derived directly from modality-specific boundary predictions. 
Wang et al.\cite{wang2024avt2} introduced dynamic cross-modal attention mechanisms to adaptively integrate single-modal classification tokens, effectively highlighting modality-specific strengths. 
Liu et al.~\cite{liu2023magnifying} further enriched dynamic fusion by proposing a tri-branch architecture (audio, video, multi-scale frames), each branch employing a Forgery Clues Magnification Transformer to amplify subtle intra-modal forgery signals. 
Subsequently, these features are adaptively integrated using learned weights based on inter-modal interactions, significantly improving sensitivity to nuanced multi-modal artifacts.

More sophisticated strategies, including attention mechanisms and MLP mixers, have recently gained traction. 
Sree et al.\cite{sree2023mis} introduced a self-attention layer applied to concatenated features, enhancing cross-modal interactions for improved detection. 
Zou et al.\cite{zou2024cross} further expanded this by incorporating multiple Transformer layers for thorough feature fusion, effectively capturing complex inter-modal relationships. 
Additionally, Raza et al.~\cite{raza2023multimodaltrace} employed an inter-modal MLP mixer layer, promoting detailed intra- and inter-modal feature interactions for enhanced integration and detection accuracy.
Complementing this research line, Wang et al.\cite{wang2024avt2} developed an AV-Local Global Interaction module to efficiently model fine-grained cross-modal correlations, improving fusion effectiveness.

Distinct from previous fusion-based strategies, Yin et al.~\cite{yin2024fine_ijcv24} addressed multi-modal deepfake detection by converting modality-specific features into a heterogeneous graph, explicitly modeling cross-modal interactions using a cross-modal graph interaction module. 
This graph-based approach effectively mines complementary audio-visual information, significantly enhancing feature integration by explicitly capturing intrinsic relationships between modalities.

\noindent \textit{Multi-task}: 
Multi-task learning has recently emerged as an effective strategy for enhancing audio-visual deepfake detection by incorporating additional tasks to explicitly guide modality-specific feature extraction, thereby improving the robustness and granularity of the detection process. 
Early methods introduced auxiliary single-modal detection tasks, enabling models to discriminate not only multi-modal but also single-modal manipulations~\cite{zhou2021joint,10081373}. 
For instance, Muppalla et al.~\cite{muppalla2023integrating} refined the granularity further by formulating a four-class classification task (real, visual-only, audio-only, and both manipulated), explicitly addressing potential variations in modality-specific manipulations.
Extending this line of research, Raza et al.~\cite{raza2023multimodaltrace} approached the problem by employing multi-task learning coupled with a cross-modal MLP mixer layer, integrating modality-specific branches and enhancing inter-modal feature interactions through multi-task supervision.
{Beyond single-modal detection as auxiliary, Zhang et al.\cite{zhang2024mfms_MM2024} incorporated a reconstructive objective to capture more robust representations on real samples.}
Further distinguishing from earlier works primarily focused on detection granularity, Yu et al.~\cite{yu2024explicit_icme2024} introduced an auxiliary correlation distillation task. 
In their method, the detection model explicitly learns content-based cross-modal correlations distilled from well-trained speech recognition models. 
This strategy effectively leverages intrinsic audio-visual semantic alignment, thereby mitigating overfitting and significantly boosting  performance.


\noindent {\textit{Advanced Representation:}}
Recent approaches enhance multi-modal deepfake detection by explicitly refining the discriminative quality of learned representations. 
Techniques to achieve this enhancement primarily include applying contrastive learning-based regularization~\cite{10081373,liu2023magnifying,zou2024cross}, sophisticated feature transformation strategies~\cite{liu2023mcl_tcsvt2023}, and disentanglement of modality-specific/invariant features~\cite{sree2023mis}.

Specifically, Yang et al.\cite{10081373} combined contrastive loss with an additive angular margin loss to simultaneously reinforce multi-modal feature separability and cross-modal alignment. 
Liu et al.\cite{liu2023magnifying} introduced a Forgery Clues Magnification Transformer, further leveraging a Jensen-Shannon divergence-based contrastive regularization to highlight subtle forgery artifacts effectively. 
Similarly, Zou et al.\cite{zou2024cross} employed contrastive learning to ensure discriminative modality-specific features while enhancing cross-modal fusion effectiveness. 
Moreover, Liu et al.\cite{liu2023mcl_tcsvt2023} proposed the Multi-modal Contrastive Learning framework, capturing fine-grained artifact cues through carefully structured multimodal contrastive objectives. 
Additionally, Mis-AvoiDD~\cite{sree2023mis} employed a unique disentanglement mechanism to separate learned representations into orthogonal modal-invariant and modal-specific subspaces, effectively reducing modality gaps and improving final  performance.


\noindent\textbf{Matching-based Learning} 
Matching-based learning leverages natural audio-visual synchronization as intrinsic cues to distinguish genuine from manipulated multi-modal media.
Early efforts by Cheng et al. \cite{cheng2023voice} focused on exploiting intrinsic voice-face correspondence by optimizing voice and face encoders using contrastive InfoNCE loss, explicitly maximizing alignment for genuine pairs and minimizing it for manipulated pairs.
Similarly, Tian et al.\cite{tian2023unsupervised} quantified cross-modal consistency through a carefully designed matching score to distinguish authentic from synthetic ones.

Recognizing critical limitations in existing benchmarks, such as subtle dataset biases, recent methods have introduced advanced matching criteria to improve robustness.
Marcella et al.\cite{Marcella_icip2024} assessed cross-modal consistency by measuring first-order statistical divergence between audio and visual feature distributions, thus capturing subtle inconsistencies indicative of manipulation.
Boldisor et al.\cite{boldisor2024circumventing_arxiv2024} proposed an audio-focused alignment network trained on pristine data to compute misalignment scores explicitly, addressing issues like artificial leading silence segments found in common benchmarks~\cite{khalid2021fakeavceleb,cai2023av}, thereby significantly enhancing generalization.
Feng et al.~\cite{feng2023self} formulated detection as an anomaly detection task, training an audio-visual synchronization model on pristine data.
During inference, media samples exhibiting anomalously low synchronization likelihoods are identified as manipulated, avoiding explicit thresholding and thus demonstrating improved generalization across diverse manipulation methods.

\noindent\textbf{Others} 
In addition to aforementioned methods, several studies have explored alternative avenues for audio-visual deepfake detection. 
Cozzolino et al.\cite{cozzolino2023audio} proposed the POI-Forensics framework, which leverages identity consistency across audio-visual streams, explicitly focusing on identifying inconsistencies tied to the portrayed individual’s identity. 
Haq et al.\cite{haq2023multimodal} introduced an emotion-based reasoning approach, utilizing psychological insights to detect unnatural emotional transitions within and between modalities.

\ding{42} Recent research have witnessed the significant evolution of audio-visual deepfake detection, successfully handling the inherently heterogeneous input signals and capturing the inter-modal correlation to enhance detection. The learning paradigms of audio-visual deepfake detection progressively evolve to complement their predecessors. As initial attempts to address this task, independent learning methods merely consider the challenge of heterogeneous inputs, ignoring the interaction between modalities. Joint learning detection algorithms design various feature fusion strategies, such as cross-attention mechanisms and MLP mixer layers, enhancing the capability to integrate modality-specific information and capture cross-modality interaction for effective audio-visual detection. Despite joint learning methods manage to obtain the inter-modal correlation for improved detection, it still requires a large amount of high-quality deepfake datasets to develop such detectors and faces the generalization challenge. Matching-based methods, which utilize the alignment of different modalities to determine authenticity, especially those in the self-supervised manner, appear to be a solution for better generalizable audio-visual detection.

Despite aforementioned advances, there still exist challenges worth attention and further efforts. Future work should pay attention to: (1) further improve the generalization capability, for instance, combining the strength of joint learning on well-curated deepfake datasets and the self-supervised learning on numerous real media corpus; (2) fine-grained detection to recognize subtle manipulations, for instance, identifying the partially manipulated segment.

\subsubsection{Text-Visual Deepfake Detection}
Recent advancements in text-visual deepfake detection have introduced sophisticated methods to address simultaneous manipulations of textual and visual content. 
However, compared to audio-visual deepfake detection, research in the text-visual domain remains relatively sparse and is currently in an early exploratory phase, with limited publicly available benchmarks and methodologies. 
Despite this limitation, existing works have made notable contributions.

Shao et al.~\cite{shao2024detecting_tpami2024} introduced the DGM4 dataset, comprising samples manipulated by distinct techniques across image and text modalities. 
They proposed the Hierarchical Multi-modal Manipulation Reasoning Transformer (HAMMER), enhancing cross-modal semantic alignment through a manipulation-aware contrastive loss, demonstrating strong performance on this comprehensive dataset.

To more effectively capture forgery artifacts, Liu et al.\cite{liu2023unified_arxiv2023} leveraged frequency-domain analysis alongside RGB images by applying discrete wavelet transforms\cite{edwards1991discrete}. 
Their method integrates both frequency and image-domain information through specialized encoders and a unified decoder, streamlining the optimization process. 
This unified decoder employs symmetric cross-modal interaction modules to extract modality-specific forgery information, complemented by a fusion interaction module to aggregate visual and textual features comprehensively.

Wang et al.~\cite{wang2024exploiting} emphasized exploiting modality-specific features for enhanced detection and grounding of multi-modal manipulations.
Their dual-branch cross-attention framework facilitates bi-directional interaction between text and visual modalities and incorporates modality-specific fine-grained classifiers to improve manipulation classification accuracy. 

\ding{42} Given the current early stage of research, future directions in text-visual deepfake detection should prioritize: (1) constructing comprehensive, diverse, and realistic benchmarks specifically designed for text-visual manipulation scenarios; (2) developing targeted cross-modal fusion and alignment strategies tailored explicitly to the semantic complexities between text and visuals; and (3) leveraging the advanced semantic reasoning and cross-modal alignment capabilities inherent in MLLMs to potentially achieve performance improvements.

\subsubsection{Discussion: Multi-modal Deepfake Detection}
Despite the rapid advancements and diversification of multi-modal deepfake detection methodologies, several significant challenges remain inadequately addressed.
In audio-visual deepfake detection, existing joint learning and fusion strategies, although effective in capturing cross-modal correlations, continue to struggle with subtle semantic discrepancies and dataset-specific biases, limiting universal generalization.
Similarly, text-visual detection research, while promising, urgently requires more comprehensive datasets and deeper theoretical analyses to effectively manage complex, multi-dimensional manipulations.
Furthermore, unlike single-modal detection, current multi-modal frameworks have insufficiently explored adversarial robustness and trustworthiness, leaving critical vulnerabilities unaddressed.
To overcome these limitations, future research could emphasize holistic, theoretically-grounded approaches, particularly leveraging Multimodal Large Language Models  to enhance detection robustness, interpretability, and generalization across modalities. 
MLLMs' powerful reasoning capabilities and intrinsic cross-modal alignment potential make them ideally suited for developing unified frameworks that simultaneously address detection accuracy, robustness, and interpretability.

\section{Challenges and Future Directions}\label{sec:future}
Facial deepfake detection has achieved considerable progress, particularly with the introduction of advanced methods across single-modal and multi-modal domains. 
Nevertheless, critical challenges persist due to the dynamic and sophisticated nature of deepfake generation techniques. 
In the following, we explore these ongoing challenges and propose potential solutions and promising research directions to effectively address them.

\subsection{Improving Cross-Scenario Generalization}
A major challenge is the limited generalization of detection methods across varying scenarios, including differences in datasets, manipulation techniques, and post-processing procedures. 
Such variations often lead to significant performance degradation when deploying models trained on one domain to another, due to notable statistical distribution shifts.
Prior research on domain adaptation~\cite{9802910_tnnls2024} and domain generalization~\cite{10420486_tnnls2024} highlights this issue but often lacks comprehensive, practical strategies tailored specifically for deepfake detection.

To address this fundamental issue effectively, future research should prioritize advanced domain adaptation methodologies. 
Adversarial-based domain alignment techniques, as explored in prior works \cite{li2021divergence_tpami2021,zhou2023self_tpami2023}, could be further adapted to align deepfake detection features across diverse generative methods and datasets. 
Additionally, meta-learning-based adaptation frameworks, inspired by recent advances \cite{vettoruzzo2024advances_tpami2024}, hold promise in enabling models to rapidly adapt to novel manipulation techniques and emerging deepfake scenarios through effective few-shot learning mechanisms.
Moreover, curriculum learning strategies, as systematically reviewed in \cite{wang2021survey_tpami2021}, may offer valuable insights by progressively introducing training complexity, thus fostering more robust generalization across various domains.
 Integrating self-supervised representation learning approaches \cite{gui2024survey_tpami2024} could also significantly enhance the robustness of feature embeddings, thereby reducing susceptibility to domain-specific artifacts and improving generalization capabilities.
Finally, leveraging ensemble learning frameworks \cite{zhou2021domain_tip2021}, which strategically combine multiple specialized models trained on distinct forgery scenarios, could yield a more resilient and comprehensive detection strategy, effectively bridging gaps across diverse real-world conditions. 
Collectively, these methodological innovations could substantially elevate the practical effectiveness and robustness of deepfake detection systems.

\subsection{Enhancing Multi-modal Alignment}
The evolution of deepfake technology has moved beyond single-modal manipulations (i.e., visual-only alterations) toward sophisticated multi-modal manipulations involving synchronized audio, visual, and textual data. 
Current single-modal detection methods become less effective in this multi-modal context due to the complexity of synchronizing and effectively integrating multiple modalities, such as audio-visual lip synchronization or text-audio semantic coherence.

To overcome this issue, future research should focus on designing advanced multi-modal fusion architectures leveraging cross-modal attention mechanisms \cite{ye2021referring_tpami2021}, contrastive multi-modal learning techniques \cite{huang2024comparison_nips2024}, and  unified multi-modal transformer \cite{wang2023unitr_iccv2023}. 
Techniques such as dynamic fusion with modality-specific attention and multi-modal prompting \cite{khattak2023maple_cvpr2023} could also significantly improve alignment and exploit correlations between modalities. 
Moreover, developing large-scale multi-modal datasets and standard benchmarks specifically designed to facilitate multi-modal analysis and evaluation would further accelerate progress and enhance comparability of methods.

\subsection{Leveraging Large Language and Vision-Language Models}
The emergence of Large Language Models (LLMs) and Vision-Language Models (VLMs) presents a promising new avenue for enhancing both the accuracy and explainability of deepfake detection systems. 
LLMs, renowned for their robust semantic reasoning and contextual understanding, offer opportunities to complement visual detection techniques by not only uncovering subtle semantic inconsistencies but also significantly enhancing detection explainability by generating detailed textual explanations that clearly articulate the rationale behind detection decisions.
Simultaneously, VLMs, exemplified by models such as CLIP and GPT-4V, demonstrate exceptional capabilities in cross-modal alignment and multimodal reasoning. 
Their ability to integrate textual descriptions and visual features makes them particularly suited to detect deepfakes by effectively correlating visual manipulations with linguistic or semantic cues. 
Despite their potential, the integration of LLMs and VLMs into deepfake detection is still nascent and faces challenges such as ensuring robust alignment between different modalities, optimizing multimodal fusion strategies, and designing precise prompting methodologies tailored explicitly for deepfake detection.

Future research should focus on exploiting the complementary strengths of LLMs and VLMs through advanced multimodal prompting strategies, structured semantic prompts, and knowledge-guided prompting techniques. Additionally, efforts should explore methods that leverage these models’ reasoning capabilities to enhance interpretability, providing clear textual explanations detailing exactly why a given media sample is identified as manipulated. These explanations would increase trustworthiness and practical usability in real-world scenarios.

\subsection{Proactive Defenses Tailored for Emerging Models}
Recent advances in generative models, particularly Diffusion Models (DMs), have elevated deepfake quality and complexity, presenting substantial challenges to existing proactive defense approaches, traditionally designed for GAN-based manipulations. 
The unique generative processes inherent in DMs, involving iterative denoising and latent space operations, necessitate specialized proactive defense strategies tailored explicitly to disrupt or watermark diffusion-generated content effectively.

Future proactive defense research should prioritize developing methods explicitly designed for DMs, focusing on disrupting critical internal processes such as latent encoding, conditional attention mechanisms, and cross-attention feature extractors. 
Approaches similar to FaceShield~\cite{jeong2024faceshield_arxiv2024}, which strategically inject minimal adversarial noise robust against common image processing operations, or latent-space watermarking methods~\cite{fernandez2023stable_iccv2023}, highlight promising directions. Additionally, investigating proactive strategies integrating cross-model robustness, provenance tracking, and identity disruption could provide comprehensive protection against emerging diffusion-based deepfake threats.

\section{Conclusion} \label{sec:conclusion}
In this survey, we comprehensively reviewed the latest developments in facial deepfake detection, particularly highlighting the state-of-the-art methodologies introduced over the past three years. 
We first outlined the scope and taxonomy of deepfake detection, followed by an overview of recent benchmark datasets and standardized evaluation metrics essential for method assessment. 
Subsequently, we systematically categorized facial deepfake detection approaches into single-modal methods that cover passive detection, specialized detection for diffusion models, and multi-modal techniques that integrate information across multiple modalities. 
Furthermore, proactive defenses designed to preemptively counter deepfake generation for both GAN-based and diffusion-based generative models were discussed in depth.
Finally, we identified critical open challenges, including the limited cross-scenario generalization, the complexity of multi-modal alignment, effective integration and utilization of large language and vision-language models, and specialized proactive defenses tailored to emerging generative technologies. 
By discussing these challenges, we outlined potential research directions aimed at building more robust, generalizable, and interpretable deepfake detection systems. 
We hope this survey provides valuable insights and facilitates further advancements in combating the evolving threats posed by increasingly sophisticated deepfakes.

\vspace{-0.5cm}

\ifCLASSOPTIONcaptionsoff
  \newpage
\fi



%



{\footnotesize
\bibliographystyle{IEEEtran}
\bibliography{mainref.bib}
}
\begin{IEEEbiography}[{\includegraphics[width=1in,height=1.25in,clip,keepaspectratio]{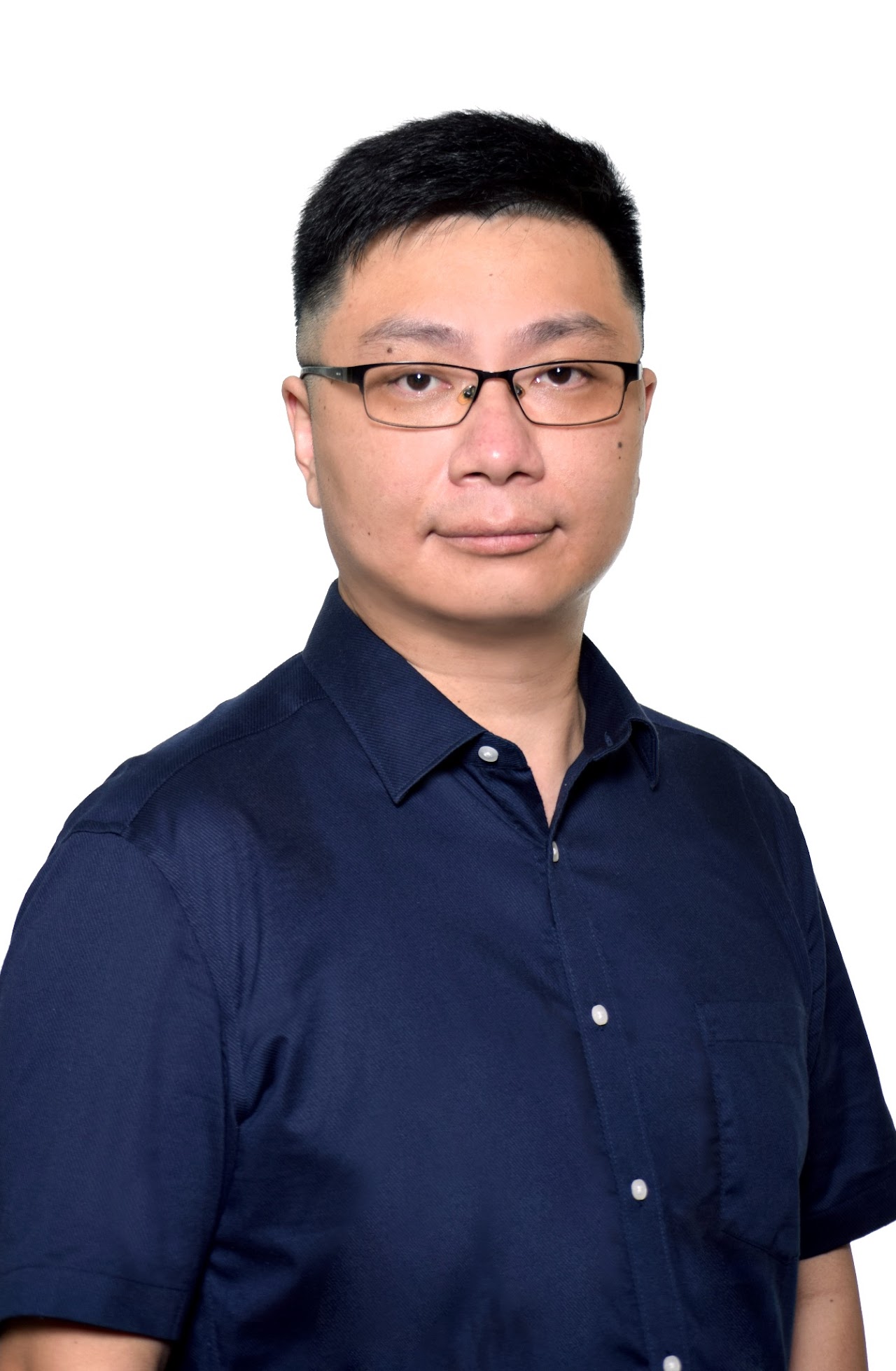}}]{Ping Liu} is an Assistant Professor in Computer Science department, University of Nevada, Reno, USA. He was a senior scientist at Centre for Frontier AI Research (CFAR), Agency for Science, Technology and Research (A*STAR), Singapore from 2020 to 2024.  From 2018 to 2020, he was a Research Staff with the Center for Artificial Intelligence, University of Technology Sydney, Ultimo, NSW, Australia.  He received the bachelor’s degree in electrical engineering from the Wuhan University of Technology, Wuhan, China, in 2005, the master’s degree from the Huazhong University of Science and Technology, Wuhan, in 2008, and the Ph.D. degree in computer science and engineering from the University of South Carolina, Columbia, SC, USA, in 2015. His research interests include computer vision and deep learning.
\end{IEEEbiography}
\begin{IEEEbiography}[{\includegraphics[width=1in,height=1.25in,clip,keepaspectratio]{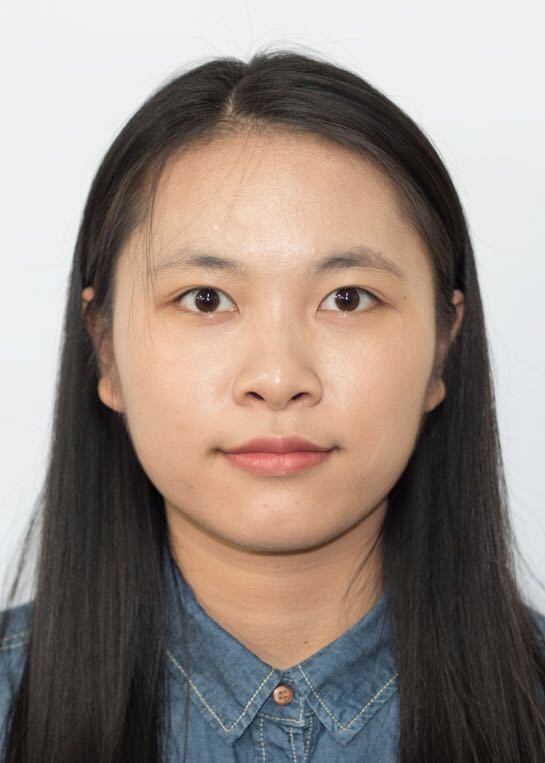}}]{Qiqi Tao} is a PhD student at Singapore University of Technology and Design (SUTD), Singapore. She obtained M.S. degree in statistics from National University of Singapore in 2024 and the B.Econ. degree in statistics from Shanghai University of Finance of Economics, Shanghai, China, in 2022. She was a research intern at the A*STAR Centre for Frontier AI Research (CFAR). Her research interests include trustworthy machine learning and AI safety, especially for Large Multi-modal Models.
\end{IEEEbiography}
\begin{IEEEbiography}[{\includegraphics[width=1in,height=1.25in,clip,keepaspectratio]{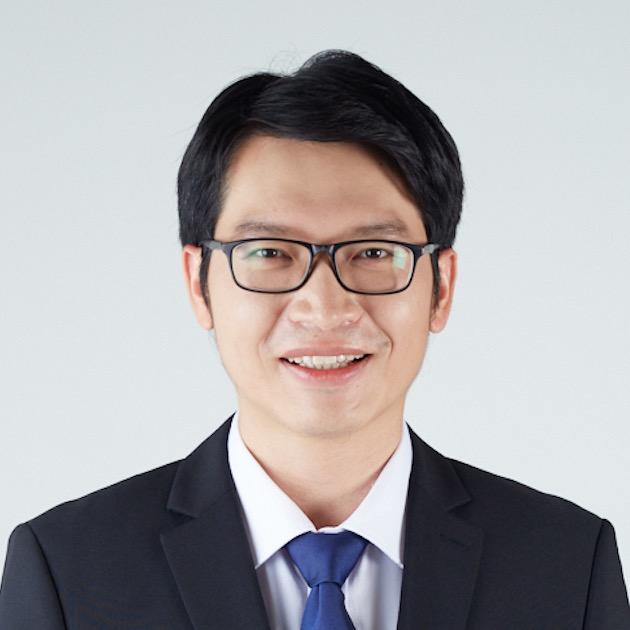}}]{Joey Tianyi Zhou} is the Deputy Director and Principal Scientist, with the A*STAR Centre for
Frontier AI Research (CFAR), Singapore. He is also holding a joint appointment with the Centre
for Advanced Technologies in Online Safety (CATOS) as principal scientist. Before working at
CFAR, he was a senior research engineer with SONY US Research Center in San Jose, USA. Dr.
Zhou received a Ph.D. degree in computer science from Nanyang Technological University (NTU),
Singapore. His current interests mainly focus on improving the efficiency and robustness of machine
learning algorithms. In these areas, he has published more than 100 papers and received the Best
Student Paper Nomination at the European Conference on Computer Vision (ECCV’16), Best
Paper Award at IEEE SmartCity 2022, International Joint Conference on Artificial Intelligence
(IJCAI) workshops, and Best Poster Award and runner-up prize at International Conference on
Computer Vision (ICCV19) on HANDS workshop and its competition, respectively.
Dr. Zhou regularly organizes workshops/tutorials at top-tier international conferences like CVPR,
IJCAI, ICDCS, etc. He is serving on an Editorial Board for many leading journals like AIJ, IEEE
Transactions, etc., and Area Chairs in top machine learning conferences like ICLR, ICML, NeurIPS,
IJCAI etc. He is listed in the Top $2\%$ Scientists Worldwide 2023 by Stanford University.
\end{IEEEbiography}
\end{document}